\documentclass{article}
\usepackage{iclr2027_conference,times}
\usepackage[hidelinks]{hyperref}
\usepackage{bbding}
\usepackage[utf8]{inputenc} 
\usepackage[T1]{fontenc}    
\usepackage{url}            
\usepackage{booktabs}       
\usepackage{amsfonts}       
\usepackage{nicefrac}       
\usepackage{microtype}      
\usepackage{xcolor}         
\usepackage{graphicx}
\usepackage{multirow}
\usepackage{array}
\usepackage{colortbl}
\usepackage{enumitem}
\usepackage{amsmath}
\usepackage{amssymb}
\usepackage{wrapfig}
\usepackage{mathtools}


\let\cite\citep

\iclrfinalcopy 

\title{Inline Memory Meets Reusable Skills: Memory-centric Framework for Vision-Language-Action Model}

\author{%
  \parbox{\dimexpr\textwidth-2\tabcolsep\relax}{%
    \centering
    \normalfont\normalsize
    \textbf{Zaijing Li}\textsuperscript{1,2}%
    \quad
    \textbf{Rui Shao}\textsuperscript{1}%
    \quad
    \textbf{Bing Hu}\textsuperscript{1}%
    \\[4pt]
    \textbf{Haoyu Zhang}\textsuperscript{1,2}%
    \quad
    \textbf{Dongmei Jiang}\textsuperscript{2}%
    \quad
    \textbf{Liqiang Nie}\textsuperscript{1}%
    \\[10pt]
    \textsuperscript{1}Harbin Institute of Technology (Shenzhen)%
    \\[2pt]
    \textsuperscript{2}Pengcheng Laboratory\\
  }%
}

\begin{document}

\maketitle
\lhead{Preprint}
\begin{abstract}
Vision-Language-Action (VLA) models have shown strong promise for general-purpose robotic manipulation, yet adapting them to new tasks and domains remains inefficient: existing methods often rely on parameter tuning, incurring substantial costs and risking catastrophic forgetting of previously learned tasks. To address this, we propose \textbf{Optimus-R}, a memory-centric VLA framework that formulates robotic adaptation as explicit query-skill memory tuning. Optimus-R introduces: (i) An \textbf{Inline Memory Interface for skill extraction}. It inserts learnable memory tokens into the VLA prefix stream, allowing the backbone to derive control-aware query and skill representations within the native action-conditioning pathway. (ii) A \textbf{Query-Skill Memory Bank for skill learning}. It externalizes skills into query prototypes for deciding \emph{what} to retrieve and skill values for specifying \emph{how} to act, supporting skill reuse and expansion with limited parameter updates. (iii) A lightweight \textbf{Bridge-and-Adapt mechanism for skill updating}. It aligns target-domain queries and skills with the existing memory space through a lightweight adapter and residual memory updates. Experiments on in-domain adaptation, cross-domain adaptation, and lifelong learning show that Optimus-R enables data-efficient skill learning while mitigating catastrophic forgetting.
\end{abstract}


\section{Introduction}
\label{sec:introduction}

\begin{wrapfigure}{r}{0.65\textwidth}
    \centering
    \small
    \vspace{-20pt}
    \includegraphics[width=0.65\textwidth]{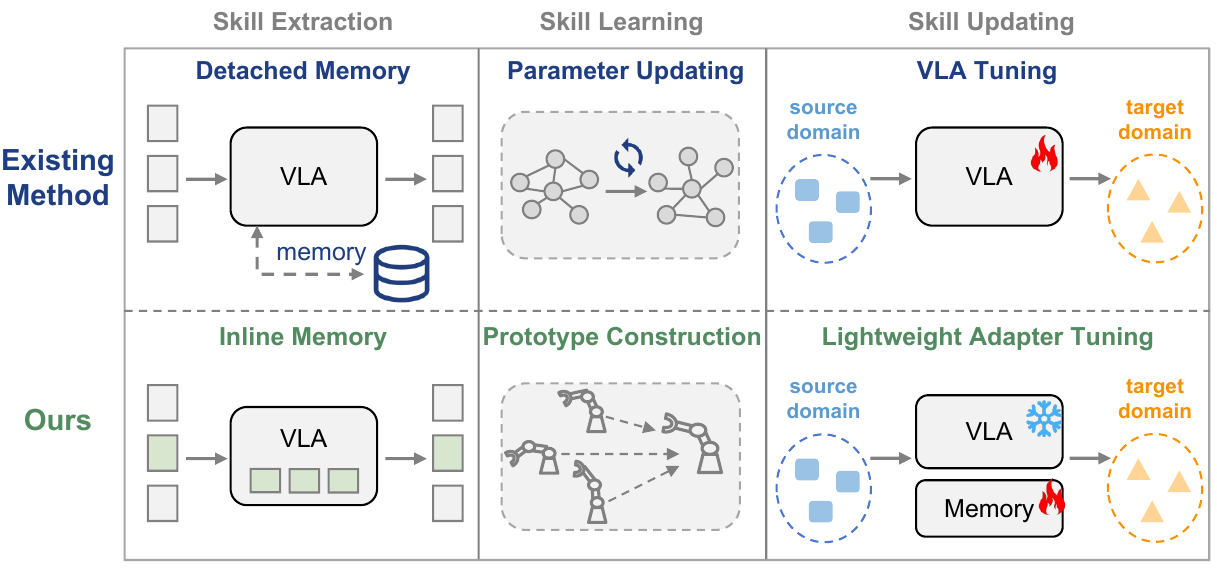}
    \caption{Comparison between parameter-centric adaptation and our memory-centric adaptation framework.}
    \label{fig:fig1}
    \vspace{-5pt}
\end{wrapfigure}

Vision-Language-Action (VLA) models~\cite{black2024pi0,kim2024openvla,kim2025openvla-oft,zhang2025dreamvla,shi2025memoryvla} have become a promising foundation for general-purpose robotic manipulation~\cite{chi2025diffusionpolicy,jiang2022vima}, benefiting from large-scale pretraining on diverse robotic datasets~\cite{liu2023libero,khazatsky2024droid,mees2022calvin,chen2025robotwin2}. However, deployment in the real world~\cite{zitkovich2023rt2,bjorck2025gr00t,intelligence2025pi06} requires robots to adapt to new tasks, objects, scenes, and domains with limited demonstrations. Most current VLAs perform such adaptation by updating model parameters~\cite{hu2021lora,kim2025openvla-oft,wen2025tinyvla,wang2026vla-adapter}. While effective, this parameter-centric paradigm makes every new skill an implicit change to the model weights, leading to high adaptation cost and possible interference with previously learned tasks. This motivates a crucial question: can a VLA acquire, store, and reuse skills through an explicit memory interface, rather than repeatedly updating its model parameters?

As shown in Fig.~\ref{fig:fig1}, a memory-centric VLA must address three coupled issues. First, skills should be derived from the VLA model itself rather than attached as detached context: external retrieval memories are only weakly coupled with the VLA action-conditioning pathway, which limits their ability to encode control-relevant information~\cite{li2026optimusvla,temiraliev2026retrieval}. Second, skills should be organized explicitly rather than absorbed into model weights: parameter updates make newly acquired behaviors hard to retrieve, expand, or reuse, and may interfere with previous skills during continual learning~\cite{wang2026lifelong}. Third, skills should remain retrievable under domain shift: visual or dynamics changes can move target-domain observations away from the source-domain memory space, causing mismatches between queries and stored skills~\cite{xu2026morphology}. These issues motivate a unified view of adaptation as skill extraction, skill learning, and skill updating.

To this end, we propose \textbf{Optimus-R}, a memory-centric framework for VLA adaptation. Optimus-R introduces : \textbf{(i) Inline Memory Interface for skill extraction.} It inserts learnable memory tokens into the VLA prefix stream, allowing the backbone to derive control-aware query and skill representations within the native policy pathway. 
\textbf{(ii) Query-Skill Memory Bank for skill learning.} It externalizes skills into query prototypes for deciding \emph{what} to retrieve and skill values for specifying \emph{how} to act, enabling new tasks to be learned by reusing, updating, or appending memory entries rather than repeatedly tuning model parameters. 
\textbf{(iii) Bridge-and-Adapt mechanism for skill updating.} It aligns target-domain queries and skill values to the existing memory space through lightweight adapter and residual memory updates, improving skill retrieval under domain shift.

We conduct experiments on in-domain adaptation, cross-domain adaptation, and in-domain lifelong learning. On LIBERO, CALVIN, and RoboTwin 2.0, Optimus-R consistently improves over the strong $\pi_{0.5}$~\cite{intelligence2025pi05} baseline under low-data regimes: with only 30\% training data, it improves the LIBERO average success rate by 16.5\%, the CALVIN average completion length by 0.42, and the RoboTwin 2.0 Hard success rate by 5.0\%. For cross-domain transfer, Optimus-R achieves a 33.3\% real-world success rate with only 20 demonstrations per task, outperforming $\pi_{0.5}$ by 13.9\%. In lifelong learning, Optimus-R improves 20-demo new-task success by 11.0\% over $\pi_{0.5}$, and reduces average forgetting from 15.0\% to 10.0\% after 80-demo adaptation.

Our main contributions are summarized as follows:
\begin{itemize}
    \item We propose \textbf{Optimus-R}, a memory-centric VLA framework that treats robotic adaptation as explicit query-skill memory tuning, reducing the need for repeated model weight updates during continual learning.
    
    \item We introduce an \textbf{Inline Memory Interface} that inserts learnable memory tokens into the VLA prefix stream, enabling the backbone to produce control-aware query and skill representations within the policy's native action-conditioning pathway.
    
    \item We design a \textbf{Query-Skill Memory Bank} that externalizes skills as decoupled query prototypes and skill values. This bank-centered design supports skill reuse, skill expansion, and continual task adaptation with limited parameter updates.
    
    \item We develop a \textbf{Bridge-and-Adapt} strategy that performs lightweight memory-space alignment for cross-domain transfer, improving the retrievability of source-domain skills under target-domain shifts such as sim-to-real adaptation.
\end{itemize}

\section{Related Work}
\label{sec:related_work}

\paragraph{Vision-Language-Action Models.}
Vision-Language-Action models \cite{zhang2025dreamvla,kim2024openvla,pertsch2025pi0fast,qu2025spatialvla,bu2025univla,song2025reconvla,song2025pdvla,li2025cogvla,li2023roboflamingo,liu2025hybridvla} are a central paradigm for robotic manipulation, spanning early multimodal policies and large generalist robot models pretrained on diverse datasets~\cite{liu2023libero,mees2022calvin,chen2025robotwin2,zitkovich2023rt2,intelligence2025pi05}. Recent work improves VLA policies through continuous action modeling \cite{li2024cogact,kim2025openvla-oft}, diffusion or flow-based generation \cite{intelligence2025pi05,chi2025diffusionpolicy}, action tokenization \cite{pertsch2025pi0fast}, and efficient adaptation~\cite{wen2025tinyvla,wang2026vla-adapter}. However, most downstream adaptation remains parameter-centric, learning new skills through model fine-tuning. Optimus-R keeps the pretrained VLA largely reusable, adapting through an explicit query-skill memory interface.

\paragraph{Memory for Embodied Agents.}
Memory mechanisms extend context and support decision making in embodied agents~\cite{zhang2024survey,he2024ma,song2024moviechat,li2025optimus2,zhu2024retrieval,xie2024embodied,li2024optimus}. In robotics, recent memory-augmented methods store demonstrations, experiences, scene histories, or external knowledge to improve long-horizon planning and manipulation~\cite{shi2025memoryvla,li2025map,lin2025echovla,li2026optimusvla}. These methods regard memory as external context or episodic retrieval. In contrast, Optimus-R injects memory into the VLA prefix stream and maintains a persistent query-skill bank for efficient adaptation.

\paragraph{Few-Shot and Continual Adaptation.}
Foundation models adapt via in-context learning~\cite{brown2020language}, prompt tuning~\cite{lester2021power}, LoRA~\cite{hu2021lora}, adapters~\cite{zhang2023llama}, or frozen-backbone transfer~\cite{li2023blip,dai2023instructblip,liu2024visual,karamcheti2024prismatic}.  Existing VLAs improve data efficiency but still encode new skills in model parameters~\cite{kim2025openvla-oft,wen2025tinyvla,wang2026vla-adapter}. Optimus-R instead externalizes skills into a global memory bank, updating only lightweight alignment modules, prototype residuals, and bank entries.
\begin{figure*}[!t]
    \centering
    \includegraphics[width=1.0\textwidth]{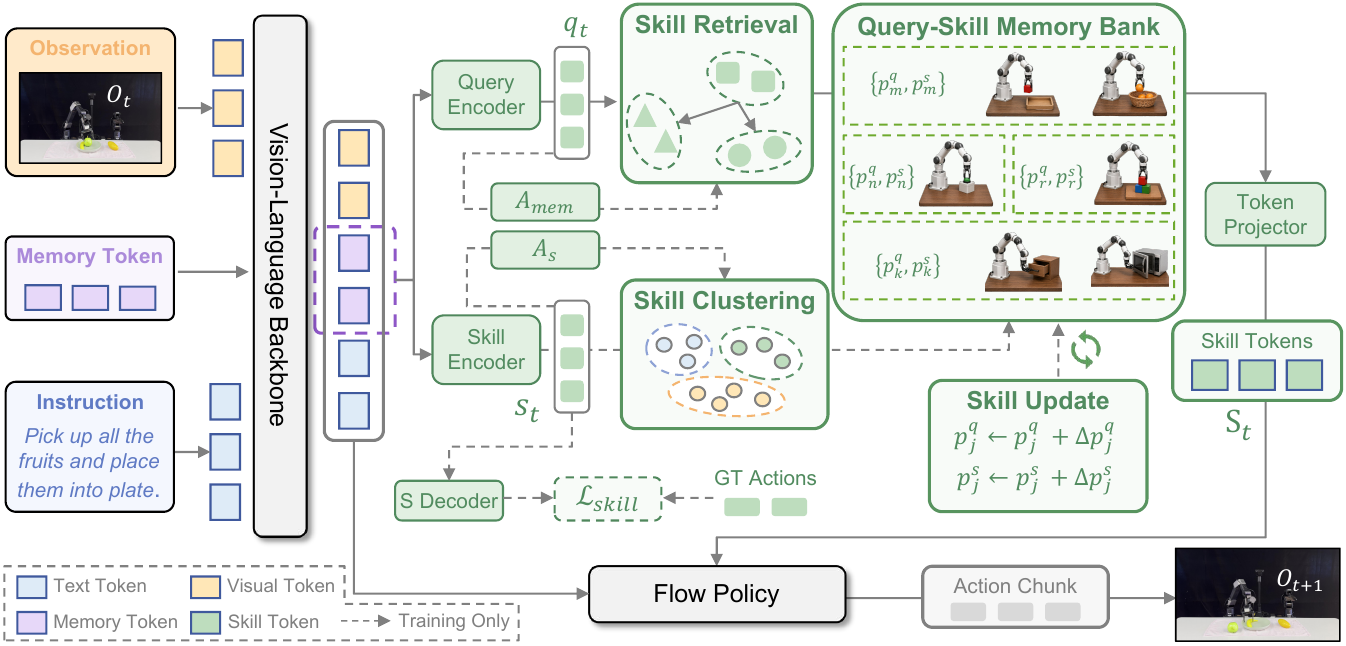}
    \caption{\textbf{Overview of Optimus-R.} Given an observation and language instruction, Optimus-R inserts learnable memory tokens into the VLA prefix stream, allowing the backbone to produce inline memory states aligned with the policy pathway. A dual-head memory readout derives a query embedding $q_t$ for retrieval and a skill embedding $s_t$ for behavior representation, with $s_t$ supervised by a skill decoder during training. The Query-Skill Memory Bank stores reusable latent skill slots as query prototypes and skill values $\{(p_j^q, p_j^s)\}$, which can be retrieved, updated, or expanded through clustering and residual memory updates. The retrieved skill is projected into VLA-compatible skill tokens and used to condition the flow policy for action-chunk generation.}
    \label{fig:fig2}
\end{figure*}

\section{Method}
\label{sec:method}

We present \textbf{Optimus-R}, a memory-centric adaptation framework for vision-language-action (VLA) models. It combines an Inline Memory Interface for extracting query and skill representations, a Query-Skill Memory Bank for storing reusable skills, and Bridge-and-Adapt training for adapting retrieval and policy conditioning from limited demonstrations.

\subsection{Overview}
\label{sec:method_overview}

As shown in Fig.~\ref{fig:fig2}, given observation $O_t$ and instruction $L$ at step $t$, the backbone $F_{\theta}$ processes multimodal input concatenated with learnable memory tokens $E^{\mathrm{mem}}$:
\begin{equation}
    [H_t^{\mathrm{vl}};M_t]
    =F_{\theta}\big(O_t,L;E^{\mathrm{mem}}\big).
    \label{eq:overview_backbone}
\end{equation}
Here, $H_t^{\mathrm{vl}}$ and $M_t$ are the updated multimodal and memory states, and $\theta$ denotes VLA parameters. Subsequently, the Query Encoder reads $M_t$ to form query $q_t$, which retrieves skill $\hat{s}_t$ from the Query-Skill Memory Bank. The retrieved skill is optionally adapted before projection into skill tokens $S_t$. Together with $H_t^{\mathrm{vl}}$, these skill tokens condition the Flow Policy $\pi_{\theta}$ to predict $\tilde{A}_t$ for the target action chunk $A_t=a_{t:t+H-1}$, where $a_t$ is an action and $H$ the horizon:
\begin{equation}
    \tilde{A}_t
    =\pi_{\theta}(H_t^{\mathrm{vl}},S_t).
    \label{eq:overview_action}
\end{equation}

\subsection{Inline Memory Interface}
\label{sec:inline_interface}

The Inline Memory Interface derives query and skill representations from the shared memory states $M_t$. To separate retrieval context from action-relevant skill content, two independent attention pooling operators read these states:
\begin{equation}
    \bar{m}_t^q=\operatorname{AttnPool}_q(M_t),
    \qquad
    \bar{m}_t^s=\operatorname{AttnPool}_s(M_t),
    \label{eq:dual_pooling}
\end{equation}
followed by a Query Encoder and a Skill Encoder:
\begin{equation}
    q_t=f_q(\bar{m}_t^q)\in\mathbb{R}^{d_q},
    \qquad
    s_t=f_s(\bar{m}_t^s)\in\mathbb{R}^{d_s}.
    \label{eq:query_skill_encode}
\end{equation}
The query embedding $q_t$ is used only for memory retrieval, and the Query Encoder supplies the retrieval query at inference. The skill embedding $s_t$ is trained to retain action-relevant information and serves as the value representation stored in the Query-Skill Memory Bank. During interface pretraining, $s_t$ is supervised by a lightweight Skill Decoder:
\begin{equation}
    \hat{A}_t^{\,s}=D_s(s_t),
    \label{eq:skill_decoder}
\end{equation}
which reconstructs the future action chunk. This auxiliary objective grounds the skill space in executable behavior. For policy conditioning, the Token Projector $T_{\psi}$ maps the skill embeddings into VLA-compatible tokens. Section~\ref{sec:bridge_adapt} specifies the conditioning latent and optimization objective for each stage.

\subsection{Query-Skill Memory Bank}
\label{sec:memory_bank}

The Query-Skill Memory Bank stores the extracted skills as decoupled key-value prototypes:
\begin{equation}
    \mathcal{B}
    =
    \{(p_j^q,p_j^s,n_j,\mathcal{E}_j)\}_{j=1}^{K},
    \label{eq:memory_bank}
\end{equation}
where $p_j^q$ is a query prototype, $p_j^s$ is the corresponding skill prototype, $n_j$ is the support count, and $\mathcal{E}_j$ is a compact set of latent query--skill pairs used for replay. The initial bank is constructed from $(q_t,s_t)$ pairs extracted by the Inline Memory Interface. To reduce redundancy, Optimus-R applies motion-aware downsampling and normalized-progress stratified sampling before clustering. It then performs Skill Clustering by first grouping samples in query space and further splitting groups whose skill variance is high. This produces prototypes that are both retrievable by state-task context and coherent in action space.

For retrieval, the current query is first aligned to the memory coordinate system:
\begin{equation}
    q'_t = A_{\mathrm{mem}}q_t.
    \label{eq:query_alignment}
\end{equation}
Let $\mathcal{N}_t$ be the top-$K_r$ prototypes under similarity to $q'_t$. Optimus-R computes the retrieval weights as
\begin{equation}
    \alpha_{tj}
    =
    \frac{
        \exp\left(
        \operatorname{sim}(q'_t,p_j^q+\Delta p_j^q)/\tau_{\mathrm{mem}}
        \right)
    }{
        \sum_{\ell\in\mathcal{N}_t}
        \exp\left(
        \operatorname{sim}(q'_t,p_{\ell}^q+\Delta p_{\ell}^q)/\tau_{\mathrm{mem}}
        \right)
    },
    \quad j\in\mathcal{N}_t,
    \label{eq:retrieval_weight}
\end{equation}
and aggregates the retrieved skill by
\begin{equation}
    \hat{s}_t
    =
    \sum_{j\in\mathcal{N}_t}
    \alpha_{tj}(p_j^s+\Delta p_j^s).
    \label{eq:retrieved_skill}
\end{equation}
The residuals $\Delta p_j^q$ and $\Delta p_j^s$ implement Skill Update without overwriting the base prototypes. Thus, adaptation changes the local memory coordinates while preserving the reusable skills.

\begin{table*}[!t]
\centering
\caption{Performance comparison on LIBERO~\cite{liu2023libero}, CALVIN~\cite{mees2022calvin}, and RoboTwin 2.0~\cite{chen2025robotwin2}. 
We report the average success rate on each LIBERO task suite, the average completion length (Avg.\ Len) on CALVIN (ABC $\rightarrow$ D), and the success rate under the Hard setting on RoboTwin 2.0. 
Data denotes the training-data ratio. $^{\dag}$ denotes reproduced results.}
\label{tb:in_domain}
\small
\setlength{\tabcolsep}{3.2pt}
\resizebox{0.90\textwidth}{!}{%
\begin{tabular}{l c | ccccc | c | c}
\toprule[1.2pt]
\multirow{2}{*}{\textbf{Method}} 
& \multirow{2}{*}{\textbf{Data}}
& \multicolumn{5}{c|}{\textbf{LIBERO}} 
& \multicolumn{1}{c|}{\textbf{CALVIN}} 
& \multicolumn{1}{c}{\textbf{RoboTwin 2.0}} \\
\cmidrule(lr){3-7} \cmidrule(lr){8-8} \cmidrule(lr){9-9}
& 
& Spatial & Object & Goal & Long & Avg.
& Avg.\ Len
& Hard \\
\midrule

DP~\cite{chi2025diffusionpolicy} 
& 100\% 
& 78.3 & 92.5 & 68.3 & 50.5 & 72.4
& 0.56
& 1.6 \\

ACT~\cite{zhao2023act} 
& 100\% 
& -- & -- & -- & -- & --
& --
& 3.5 \\

DP3~\cite{ze2024dp3} 
& 100\% 
& -- & -- & -- & -- & --
& --
& 5.2 \\

RDT~\cite{liu2024rdt} 
& 100\% 
& -- & -- & -- & -- & --
& --
& 18.4 \\

MemoryVLA~\cite{shi2025memoryvla} 
& 100\% 
& 98.4 & 98.4 & 96.4 & 93.4 & 96.7
& --
& -- \\

OpenVLA-OFT~\cite{kim2025openvla-oft} 
& 100\% 
& 97.6 & 98.4 & 97.9 & \underline{94.5} & 97.1
& 4.10
& -- \\

DreamVLA~\cite{zhang2025dreamvla} 
& 100\% 
& 97.5 & 94.0 & 89.5 & 89.5 & 92.6
& \textbf{4.44}
& -- \\

VLA-Adapter~\cite{wang2026vla-adapter} 
& 100\% 
& 97.8 & \underline{99.2} & 97.2 & \textbf{95.0} & \underline{97.3}
& \underline{4.42}
& -- \\

ReconVLA~\cite{song2025reconvla} 
& 100\% 
& -- & -- & -- & -- & --
& 3.95
& -- \\

OpenVLA~\cite{kim2024openvla} 
& 100\% 
& 84.7 & 88.4 & 79.2 & 53.7 & 76.5
& 3.27
& -- \\

UniVLA~\cite{bu2025univla} 
& 100\% 
& 95.4 & 98.8 & 93.6 & 94.0 & 95.4
& 3.80
& -- \\

$\pi_{0}$~\cite{black2024pi0} 
& 100\% 
& 96.8 & 98.8 & 95.8 & 85.2 & 94.2
& 3.92
& 22.9 \\

\midrule

$\pi_{0.5}$$^{\dag}$~\cite{intelligence2025pi05} 
& 30\% 
& 76.6 & 75.0 & 74.8 & 62.0 & 72.1
& 2.09
& 9.8 \\

\rowcolor[HTML]{F3F6FF}
Optimus-R 
& 30\% 
& 89.6 & 91.8 & 87.0 & 86.0 & 88.6
& 2.51
& 14.8 \\

$\pi_{0.5}$$^{\dag}$~\cite{intelligence2025pi05} 
& 50\% 
& 92.8 & 94.2 & 93.2 & 88.0 & 92.1
& 3.14
& 19.6 \\

\rowcolor[HTML]{F3F6FF}
Optimus-R 
& 50\% 
& 97.0 & 97.0 & 96.8 & 91.4 & 95.6
& 3.57
& 21.6 \\

$\pi_{0.5}$$^{\dag}$~\cite{intelligence2025pi05} 
& 70\% 
& 96.2 & 95.6 & 96.8 & 91.4 & 95.0
& 3.88
& 30.1 \\

\rowcolor[HTML]{F3F6FF}
Optimus-R 
& 70\% 
& 98.2 & 97.8 & 98.0 & 92.2 & 96.6
& 3.94
& 36.4 \\

$\pi_{0.5}$$^{\dag}$~\cite{intelligence2025pi05} 
& 100\% 
& 98.8 & 98.2 & 98.0 & 92.4 & 96.9
& 4.26
& 46.6 \\

\rowcolor[HTML]{E7EEFE}
Optimus-R 
& 100\% 
& \textbf{99.2} & \textbf{99.4} & \textbf{98.4} & 94.2 & \textbf{97.8}
& 4.38
& \textbf{55.0} \\

\bottomrule[1.2pt]
\end{tabular}%
}
\end{table*}
\subsection{Bridge-and-Adapt Training}
\label{sec:bridge_adapt}

Stage A learns the interface and constructs the initial bank. Building on this bank, Stage B bridges cross-domain shifts, while Stage C supports in-domain lifelong memory adaptation.

\paragraph{Stage A: interface pretraining.}
Stage A trains the memory tokens, Query Encoder, Skill Encoder, Skill Decoder, Token Projector, and the last layers of the VLA backbone. During this stage, the Flow Policy is conditioned on skill tokens $S_t=T_{\psi}(s_t)$ projected from the encoded skill. The policy and interface are jointly optimized with
\begin{equation}
    \mathcal{L}_{\mathrm{pre}}
    =
    \mathcal{L}_{\mathrm{fm}}
    +
    \lambda_q\mathcal{L}_{q}^{\mathrm{nce}}
    +
    \lambda_s\mathcal{L}_{\mathrm{skill}},
    \qquad
    \mathcal{L}_{\mathrm{skill}}
    =
    \operatorname{Huber}(D_s(s_t),A_t).
    \label{eq:pretrain_loss}
\end{equation}
Here, $\mathcal{L}_{\mathrm{fm}}$ is the original VLA flow-matching loss. The query contrastive loss $\mathcal{L}_{q}^{\mathrm{nce}}$ uses samples from the same task and nearby normalized progress as positives, together with local neighboring states from the same trajectory. After Stage A, the Inline Memory Interface, Token Projector, Skill Decoder, and backbone are frozen, and the initial memory bank is built.

\paragraph{Stage B: bridge adaptation.}
For cross-domain streams such as sim-to-real transfer, visual, sensory, or dynamics shifts can change the query and skill representations associated with the same behavior. To accommodate these shifts, Stage B unfreezes the backbone and updates it alongside $A_{\mathrm{mem}}$, $\tau_{\mathrm{mem}}$, and active prototype residuals, while keeping the learned interface modules, Token Projector, and base memory prototypes fixed. Here, query alignment maps target-domain queries into the bank's retrieval coordinates. When the skill-space mismatch is large, a low-rank residual Skill Adapter further corrects the retrieved skill toward the current encoded behavior target:
\begin{equation}
    \tilde{s}_t
    =
    A_s(\hat{s}_t)
    =
    \hat{s}_t+U_sV_s^{\top}\hat{s}_t,
    \qquad
    U_s,V_s\in\mathbb{R}^{d_s\times r},\quad r\ll d_s.
    \label{eq:skill_adapter}
\end{equation}
The residual is zero-initialized so that the adapter initially preserves $\hat{s}_t$. The corrected skill then supplies policy tokens $S_t=T_{\psi}(\tilde{s}_t)$ in Stages B/C and deployment, with $\tilde{s}_t=\hat{s}_t$ when the adapter is inactive. During Stage B, the adapter is optimized through both policy and alignment losses; it is then kept fixed in Stage C.

\paragraph{Stage C: memory adaptation.}
For in-domain continual learning, Stage C keeps the VLA backbone and Inline Memory Interface frozen and confines adaptation to $A_{\mathrm{mem}}$, $\tau_{\mathrm{mem}}$, active prototype residuals, and the memory bank. When an incoming sample's aligned query is close to an existing prototype, Optimus-R updates the corresponding residual and support statistics. Otherwise, the sample enters a candidate buffer. A new prototype is appended only when a cluster of buffered samples has sufficient support and low query--skill variance. Further details are provided in Appendix~\ref{app:bank_update_rule}.

To preserve previously learned retrieval associations during these updates, each batch supplements current-task samples with stored latent query--skill pairs in the alignment term. These exemplars provide skill-space supervision for previously observed behaviors without requiring raw images or action trajectories. Meanwhile, the flow-matching loss uses only current-task demonstrations to supervise action generation on the new task.

Stages B and C update different components but optimize the same memory adaptation objective:
\begin{equation}
    \mathcal{L}_{\mathrm{life}}
    =
    \mathcal{L}_{\mathrm{fm}}
    +
    \lambda_{\mathrm{align}}
    \left\|
        \tilde{s}_t-\operatorname{sg}(s_t)
    \right\|_2^2
    +
    \lambda_{\mathrm{reg}}
    \sum_{j\in\mathcal{A}_t}
    \left(
        \|\Delta p_j^q\|_2^2
        +
        \|\Delta p_j^s\|_2^2
    \right),
    \label{eq:life_loss}
\end{equation}
where $\mathcal{A}_t$ denotes the active retrieved prototypes and $\operatorname{sg}(\cdot)$ stops gradients through the encoded skill target $s_t$. The flow-matching term supervises memory conditioning through action generation, while the alignment term directly supervises the retrieved skill after any adapter correction. The residual regularizer limits prototype drift during these updates.

\setlength{\textfloatsep}{14pt plus 2pt minus 2pt}
\setlength{\dbltextfloatsep}{14pt plus 2pt minus 2pt}
\setlength{\floatsep}{8pt plus 2pt minus 2pt}
\setlength{\dblfloatsep}{8pt plus 2pt minus 2pt}
\setlength{\abovecaptionskip}{8pt}

\begin{figure*}[!t]
    \centering
    \includegraphics[width=1.0\textwidth]{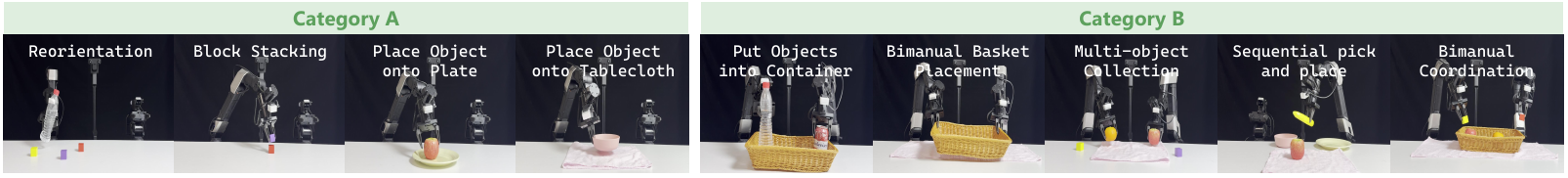}
    \caption{Overview of Real-World tasks. We classify tasks into two categories (A and B) for lifelong setting: the model first learns categories A tasks and is then incrementally adapted to categories B tasks. Details are provided in the Appendix.}
    \label{fig:fig3}
\end{figure*}
\begin{table*}[!t]
\centering
\caption{Cross-Domain evaluation across 9 Real-World tasks. Samples denotes the number of training demonstrations per task.
We report the success rate (\%) for each task. Compared to the baselines, Optimus-R demonstrates superior performance across different data scales.}
\label{tab:real_world_lifelong_results}
\small
\setlength{\tabcolsep}{3.0pt}
\resizebox{0.9\textwidth}{!}{%
\begin{tabular}{l c | cccccccccc}
\toprule[1.2pt]
\textbf{Method} 
& \textbf{Samples}
& \textbf{R.O.}
& \textbf{B.S.}
& \textbf{P.P.}
& \textbf{P.T.}
& \textbf{P.C.}
& \textbf{B.P.}
& \textbf{M.C.}
& \textbf{S.P.P.}
& \textbf{B.Sq.}
& \textbf{Avg} \\
\midrule

RDT~\cite{liu2024rdt}
& 80
& 20.0 & 40.0 & 35.0 & 50.0 & 35.0 & 30.0 & 20.0 & 5.0 & 0.0 & 26.1 \\

OpenVLA~\cite{kim2024openvla}
& 80
& 15.0 & 35.0 & 45.0 & 55.0 & 40.0 & 20.0 & 15.0 & 20.0 & 10.0 & 28.3 \\

OpenVLA-OFT~\cite{kim2025openvla-oft}
& 80
& 65.0 & 75.0 & 80.0 & 70.0 & 55.0 & 45.0 & 50.0 & 20.0 & 30.0 & 54.4 \\

$\pi_{0}$~\cite{black2024pi0}
& 80
& 75.0 & 75.0 & 80.0 & 85.0 & 45.0 & 55.0 & 50.0 & 30.0 & 20.0 & 57.2 \\

\midrule

$\pi_{0.5}$$^{\dag}$~\cite{intelligence2025pi05}
& 20
& 25.0 & 20.0 & 35.0 & 25.0 & 30.0 & 25.0 & 10.0 & 5.0 & 0.0 & 19.4 \\

\rowcolor[HTML]{F3F6FF}
Optimus-R
& 20
& 40.0 & 40.0 & 55.0 & 40.0 & 35.0 & 35.0 & 30.0 & 10.0 & 15.0 & 33.3 \\

$\pi_{0.5}$$^{\dag}$~\cite{intelligence2025pi05}
& 50
& 65.0 & 55.0 & 70.0 & 65.0 & 40.0 & 40.0 & 30.0 & 20.0 & 15.0 & 44.4 \\

\rowcolor[HTML]{F3F6FF}
Optimus-R
& 50
& 75.0 & 70.0 & 80.0 & 80.0 & 60.0 & 65.0 & 50.0 & 40.0 & 35.0 & 61.7 \\

$\pi_{0.5}$$^{\dag}$~\cite{intelligence2025pi05}
& 80
& 75.0 & 75.0 & 80.0 & 85.0 & 60.0 & 60.0 & 50.0 & 45.0 & 40.0 & 63.3 \\

\rowcolor[HTML]{E7EEFE}
Optimus-R
& 80
& \textbf{85.0} & \textbf{90.0} & \textbf{90.0} & \textbf{85.0} 
& \textbf{70.0} & \textbf{65.0} & \textbf{60.0} & \textbf{55.0} & \textbf{50.0} & \textbf{72.2} \\

\bottomrule[1.2pt]
\end{tabular}%
}
\end{table*}

\section{Experiments}
\label{sec:experiments}

We evaluate Optimus-R through three research questions:
\textbf{Q1}. Does Optimus-R improve sample efficiency on diverse simulation benchmarks?
\textbf{Q2}. Can Optimus-R adapt simulation-acquired skills to the real-world domain with limited demonstrations?
\textbf{Q3}. Can Optimus-R acquire new real-world skills while preserving previously learned ones?

\subsection{Implementation Details}
\label{subsec:implementation}

We initialize Optimus-R from pretrained $\pi_{0.5}$~\cite{intelligence2025pi05} and add memory tokens, the query-skill interface, and the skill bank. We use $m=4$ memory tokens, dimensions $d_q=d_s=256$, $n_s=4$ skill tokens per retrieved prototype, top-$2$ retrieval, and an initial bank of $K_0=128$ prototypes. Training uses 8$\times$ NVIDIA A800 GPUs, a global batch size of 256, and 30,000 steps. Further details are in the Appendix.

\subsection{In-Domain Adaptation}
\label{subsec:simulation}

\paragraph{Experimental Setup.}
We evaluate sample efficiency on LIBERO~\cite{liu2023libero}, CALVIN~\cite{mees2022calvin}, and RoboTwin 2.0~\cite{chen2025robotwin2}, adapting the same Stage-A pretrained model with 30\%, 50\%, 70\%, and 100\% of the demonstrations. We report average success over the four LIBERO suites (Spatial, Object, Goal, and Long; 500 rollouts per suite), average completed sequence length on CALVIN ABC $\rightarrow$ D (500 rollouts), and RoboTwin 2.0 \textit{Hard} success rates (100 rollouts per task).

\paragraph{Results and Analysis.}
Table~\ref{tb:in_domain} shows consistent gains over $\pi_{0.5}$~\cite{intelligence2025pi05} across all three benchmarks and data regimes. With 30\% training data, Optimus-R improves LIBERO average success by 16.5 percentage points, with a particularly large gain on LIBERO-Long. This suggests that reusable skill prototypes benefit long-horizon tasks with temporally structured behaviors. With full data, Optimus-R achieves the best LIBERO average and RoboTwin 2.0 Hard performance among the compared methods, while remaining competitive on CALVIN. These results support our design goal: using target-domain demonstrations to refine and reuse externalized skill memory rather than relearning the policy through repeated global parameter updates.

\begin{figure*}[!t]
    \centering
    \includegraphics[width=1.0\textwidth]{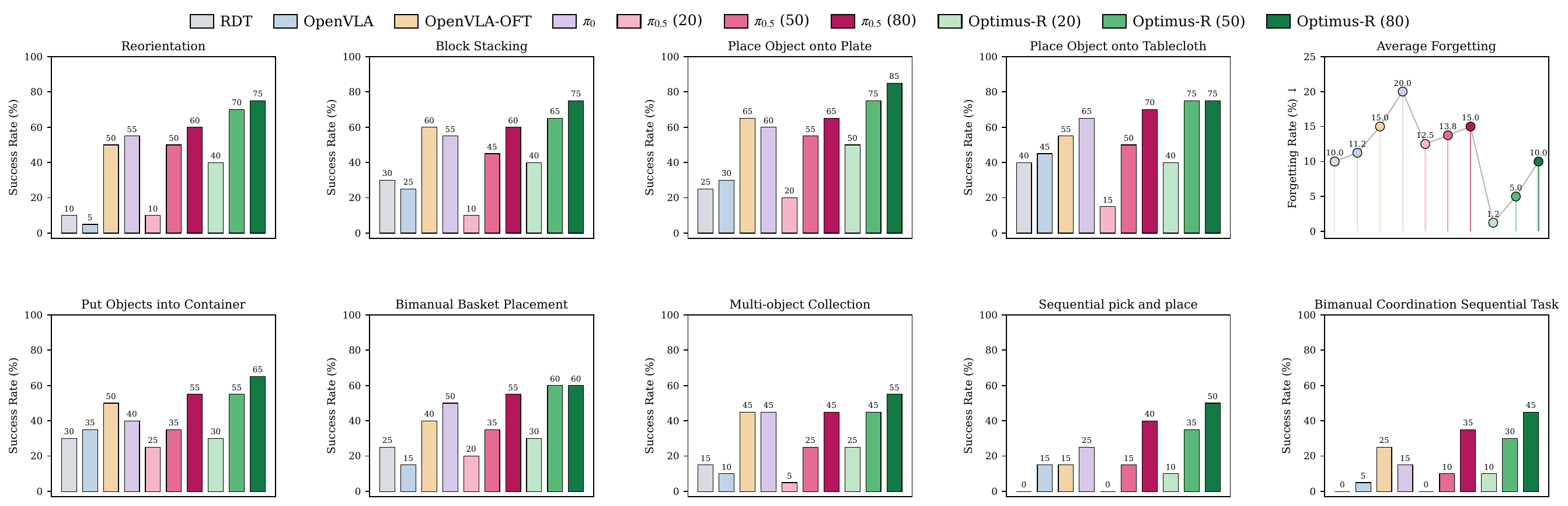}
    \caption{Real-world in-domain lifelong learning results. The model first learns $N_{\mathrm{old}}=4$ tasks and is then incrementally adapted to $N_{\mathrm{new}}=5$ tasks. The first row reports the post-adaptation success rates on the previously learned tasks, together with the average forgetting rate, where lower forgetting indicates better retention. The second row reports the success rates on the newly learned tasks after incremental adaptation. Optimus-R achieves higher new-task performance while better preserving old-task performance.}
    \label{fig:fig4}
\end{figure*}

\subsection{Cross-Domain Adaptation}
\label{subsec:sim2real}

\paragraph{Experimental Setup.}
We transfer a RoboTwin 2.0-pretrained model to nine real-world tasks with 20, 50, and 80 demonstrations per task, reporting average success over 20 rollouts per task. The platform is a 14-DoF bimanual GALAXEA R1 Lite robot with wrist-mounted and third-person cameras at $224\times224$ resolution. Figure~\ref{fig:fig3} shows the two task categories; further details are in the Appendix. This setting tests whether the bank provides reusable simulation-acquired manipulation skills that can be grounded in a new domain with limited real-world data.

\paragraph{Results and Analysis.}
Table~\ref{tab:real_world_lifelong_results} shows that Optimus-R reaches 33.3\% success with 20 demonstrations per task, exceeding $\pi_{0.5}$ by 13.9 percentage points. This indicates useful transferable skill priors under limited target-domain supervision. With 80 demonstrations, Optimus-R achieves the highest average success among the compared methods. Gains also extend to long-horizon and sequential tasks such as S.P.P. and B.Sq., where adaptation is more sensitive to visual and dynamics shifts. These results suggest that the bank preserves simulation-acquired manipulation structure, while Bridge-and-Adapt aligns target-domain queries and skill values with the existing memory space.

\subsection{In-Domain Lifelong Learning}
\label{subsec:lifelong}
 
\paragraph{Experimental Setup.}
We evaluate lifelong learning by first training on $N_{\mathrm{old}}=4$ real-world tasks (Category A), then adapting to $N_{\mathrm{new}}=5$ new tasks (Category B) under the same 20 /  50 / 80 demonstration settings. We report new-task success, old-task success after adaptation, and average forgetting as evaluation metrics:
\begin{equation}
    \mathcal{F}
    =
    \frac{1}{N_{\mathrm{old}}}
    \sum_{i=1}^{N_{\mathrm{old}}}
    \left(
    S_i^{\mathrm{before}} - S_i^{\mathrm{after}}
    \right),
\end{equation}
where $S_i^{\mathrm{before}}$ and $S_i^{\mathrm{after}}$ denote the success rate of old task $i$ before and after new-task adaptation. This setting tests whether Optimus-R can acquire new skills while preserving learned tasks.

\paragraph{Results and Analysis.}
With 20 demonstrations per new task, Optimus-R achieves 21.0\% success, exceeding $\pi_{0.5}$ by 11.0 percentage points (Fig.~\ref{fig:fig4}). After 80-demo adaptation, it retains 77.5\% old-task success, with 10.0\% forgetting versus 15.0\% for both OpenVLA-OFT and $\pi_{0.5}$. Together, these results answer \textbf{Q3}: prototype expansion and localized residual updates support acquisition of new bimanual skills while limiting interference with previously learned behaviors.

\begin{table}[!t]
\centering
\small
\caption{Adaptation cost and LIBERO performance. GPU hours correspond to the 100\% data setting. Parameter counts are rounded; M and B denote millions and billions.}
\label{tab:app_adaptation_cost}
\setlength{\tabcolsep}{3pt}
\renewcommand{\arraystretch}{1.0}
\resizebox{0.94\textwidth}{!}{%
\begin{tabular}{l|ccc|cc|cccc}
\toprule[1.2pt]
\multirow{2}{*}{\textbf{Method}}
& \multirow{2}{*}{\shortstack{\textbf{Total}\\\textbf{params}}}
& \multirow{2}{*}{\shortstack{\textbf{Added}\\\textbf{params}}}
& \multirow{2}{*}{\shortstack{\textbf{Trainable}\\\textbf{params}}}
& \multirow{2}{*}{\shortstack{\textbf{GPU}\\\textbf{hours}}}
& \multirow{2}{*}{\shortstack{\textbf{Inference}\\\textbf{Hz}}}
& \multicolumn{4}{c}{\textbf{Average SR (\%) by data ratio}} \\
\cmidrule(lr){7-10}
& & & & & & \textbf{30\%} & \textbf{50\%} & \textbf{70\%} & \textbf{100\%} \\
\midrule
$\pi_{0.5}$ (Full FT) & 3.6B & -- & 3.6B & 560 & 17 & 72.1 & 92.1 & 95.0 & 96.9 \\
$\pi_{0.5}$ (LoRA) & 3.9B & 0.3B & 0.3B & 128 & 12 & 71.5 & 86.9 & 91.2 & 96.5 \\
OpenVLA-OFT (LoRA) & 7.8B & 0.3B & 0.3B & 320 & 10 & 66.4 & 83.5 & 89.1 & 96.4 \\
\midrule
\rowcolor[HTML]{E7EEFE}
Optimus-R & 3.6B & 7M & 0.3B & 86 & 14 & \textbf{88.6} & \textbf{95.6} & \textbf{96.6} & \textbf{97.8} \\
\bottomrule[1.2pt]
\end{tabular}%
}
\end{table}

\begin{table*}[!t]
\centering
\caption{Ablation success rates (\%) on RoboTwin 2.0 (Hard) and nine real-world tasks under different data budgets. Variant definitions are given in Sec.~\ref{subsec:ablation}.}
\label{tb:ablation_study}
\small
\setlength{\tabcolsep}{8.0pt}
\resizebox{0.90\textwidth}{!}{%
\begin{tabular}{l | cc | ccc}
\toprule[1.2pt]
\multirow{2}{*}{\textbf{Model Variant}} 
& \multicolumn{2}{c|}{\textbf{RoboTwin 2.0 (In-Domain)}} 
& \multicolumn{3}{c}{\textbf{Real-World 9 Tasks (Cross-Domain)}} \\
\cmidrule(lr){2-3} \cmidrule(lr){4-6}
& 50\% Data & 100\% Data & 20 Demos & 50 Demos & 80 Demos \\
\midrule

w/o Inline Memory 
& 16.5 & 45.2 & 20.6 & 48.3 & 60.0 \\

Coupled Query-Skill 
& 18.2 & 48.0 & 22.8 & 50.0 & 62.8 \\

w/o Prototype Residuals 
& 17.0 & 46.5 & 17.8 & 46.7 & 58.3 \\

w/o Stage-B Bridge 
& 21.4 & 53.8 & 12.8 & 35.6 & 48.3 \\

\midrule
\rowcolor[HTML]{E7EEFE}
\textbf{Optimus-R (Full)} 
& \textbf{21.6} & \textbf{55.0} & \textbf{33.3} & \textbf{61.7} & \textbf{72.2} \\

\bottomrule[1.2pt]
\end{tabular}%
}
\end{table*}

\subsection{Comparison with Parameter-Efficient Baselines}
\label{app:adaptation_cost}

We compare Optimus-R with full fine-tuning and LoRA adaptation of $\pi_{0.5}$~\cite{intelligence2025pi05}, and LoRA adaptation of OpenVLA-OFT~\cite{kim2025openvla-oft}, on LIBERO at four training-data ratios. Table~\ref{tab:app_adaptation_cost} reports parameter counts, average success rates, training cost (GPU hours), and inference frequency (Hz). Training costs correspond to the 100\% data setting.

Optimus-R achieves the highest success rate at every data ratio, with the largest gains under limited data. At full data, it reaches 97.8\% success in 86 GPU hours, versus 96.5\% in 128 GPU hours for $\pi_{0.5}$-LoRA, with both reporting approximately 0.3B trainable parameters. This comparison indicates improved adaptation efficiency at a comparable trainable parameter budget. Its inference frequency of 14\,Hz exceeds both LoRA baselines (12\,Hz and 10\,Hz).

\subsection{Ablation Study}
\label{subsec:ablation}

\paragraph{Experimental Setup.}
We ablate four components on RoboTwin 2.0 and real-world tasks (Table~\ref{tb:ablation_study}): Inline Memory (replaced by detached prompts), query--skill separation (replaced by a shared latent), prototype residuals (no localized $\Delta p$ updates), and Stage-B Bridge (memory expansion without coordinate alignment). More ablation experiments are provided in Appendix.

\paragraph{Results and Analysis.}
Every ablation reduces success, supporting the joint role of control-aware memory representations and localized adaptation. The clearest domain-dependent effect comes from Stage-B Bridge: removing it has a small in-domain impact but lowers 20-demo real-world success from 33.3\% to 12.8\%. This contrast suggests that expanding memory alone is insufficient under substantial domain shift; aligning target-domain queries and skills with the existing bank is critical for transfer.

\subsection{Qualitative Analysis}
\label{subsec:qualitative}

\begin{figure*}[!t]
    \centering
    \includegraphics[width=0.95\textwidth]{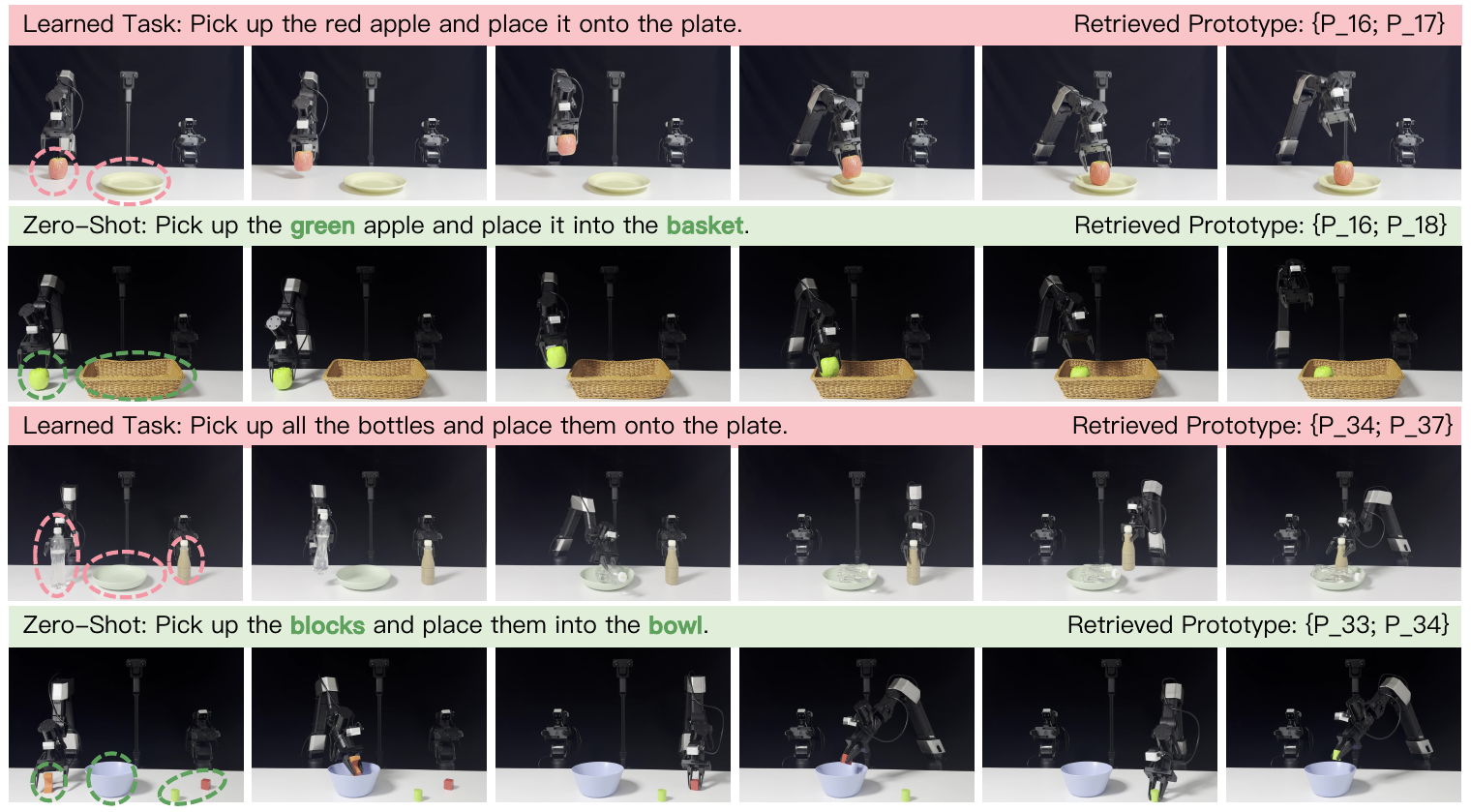}
    \caption{\textbf{Qualitative results of Optimus-R on real-world tasks.} Each row shows a rollout sequence from left to right. Rows 1 and 3 correspond to learned tasks, where Optimus-R executes previously trained pick-and-place behaviors, including placing a red apple onto a plate and placing bottles onto a plate. Rows 2 and 4 show zero-shot task variations, where the model transfers the learned manipulation structure to new object-receptacle combinations, such as placing a green apple into a basket and placing blocks into a bowl.}
    \label{fig:fig5}
\end{figure*}

Figure~\ref{fig:fig5} contrasts learned tasks with zero-shot variations in object appearance, category, receptacle, and task semantics. Successful transfer to new object--receptacle combinations suggests reuse of manipulation skills beyond fixed training pairs. Given a new instruction and observation, query representations retrieve compatible skill values from the Query-Skill Memory Bank, which are projected into VLA-compatible skill tokens to guide action generation without additional parameter updates. This behavior is consistent with our design: skills extracted through inline memory are organized as query-skill entries and reused across task variations.

\begin{samepage}
\subsection{Retrieval-Space Visualization}
\label{app:retrieval_visualization}

\begin{wrapfigure}{r}{0.50\textwidth}
    \centering
    \vspace{-20pt}
    \includegraphics[width=\linewidth]{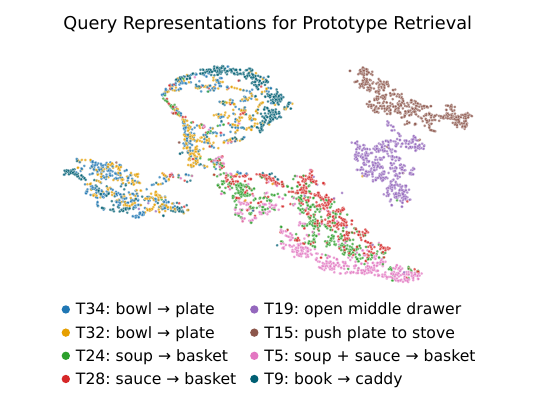}
    \caption{Aligned queries colored by task. All 4,000 queries are shown; labels such as \texttt{T34} denote task IDs. Prototype markers are omitted.}
    \vspace{-15pt}
    \label{fig:app_retrieval_visualization}
\end{wrapfigure}
We record the top-1 prototype selected during offline inference on 4,000 observations from eight LIBERO tasks, using 20 demonstrations per task and 25 progress-stratified observations per demonstration. Figure~\ref{fig:app_retrieval_visualization} visualizes the associated aligned queries $q'_t\in\mathbb{R}^{256}$, colored by task. The queries and their retrieved keys are $\ell_2$-normalized and jointly projected using t-SNE; only queries are displayed. The two bowl-to-plate tasks (\texttt{T34}, \texttt{T32}) occupy overlapping local neighborhoods, while soup-to-basket, sauce-to-basket, and multi-object collection (\texttt{T24}, \texttt{T28}, \texttt{T5}) share a broad region. Drawer opening (\texttt{T19}) and plate pushing (\texttt{T15}) form more distinct regions. This structure is consistent with related tasks sharing retrieval neighborhoods.
\par
\end{samepage}

\section{Conclusion}
\label{sec:conclusion}

We presented \textbf{Optimus-R}, a memory-centric VLA framework that formulates robotic adaptation as explicit query-skill memory tuning. Optimus-R introduces three key designs: an \textbf{Inline Memory Interface} for deriving control-aware query and skill representations within the native VLA policy pathway, a \textbf{Query-Skill Memory Bank} for externalizing skills into reusable and expandable memory entries, and a lightweight \textbf{Bridge-and-Adapt} mechanism for aligning target-domain queries and skills with the existing memory space. Experiments across in-domain adaptation, cross-domain adaptation, and lifelong learning show that Optimus-R enables data-efficient adaptation, improves skill reuse, and reduces reliance on conventional fine-tuning.
\clearpage

\subsection*{AI use statement}

We used generative AI tools solely to polish the manuscript's language and improve clarity and readability. No generative AI tools were used for other aspects of this research, including research ideation, method development, experimental design, implementation, data generation or processing, or analysis and interpretation of results. The authors take full responsibility for the final content, including all scientific claims and conclusions.

\subsection*{Reproducibility statement}
To facilitate reproducibility, we provide detailed implementation procedures and hyperparameter settings in Appendices~\ref{app:method_details} and~\ref{app:training_objectives}, covering the model architecture, stage-wise training, and memory-bank construction and update rules. Additional experimental settings and evaluation protocols are described in the appendix.
\bibliography{main}
\bibliographystyle{iclr2027_conference}
\clearpage


\appendix

\section{Broader Impacts}
Optimus-R aims to improve the data efficiency and reusability of Vision-Language-Action models by shifting robotic adaptation from repeated parameter updates to explicit query-skill memory tuning. This design may reduce the amount of task-specific data and computation required to adapt robots to new environments, thereby lowering the cost of deploying robotic systems. The memory-centric formulation also provides a more explicit interface for inspecting, reusing, and updating learned skills, which may support more maintainable lifelong robotic systems.

At the same time, improved adaptation efficiency can also increase the risk of deploying robots in insufficiently validated environments. Retrieved or reused skills may behave unexpectedly under unseen objects, unsafe scene configurations, or distribution shifts beyond those covered during adaptation. Moreover, real-world robotic data may contain sensitive visual information, and memory banks or stored exemplars should be managed with appropriate privacy and data-governance safeguards. 
\section{Limitation}
Despite its effectiveness, Optimus-R has several limitations. First, the framework relies on a strong pretrained VLA backbone; when the base policy lacks the necessary manipulation primitives, memory retrieval and prototype expansion alone may be insufficient. Second, although the Query-Skill Memory Bank improves skill reuse, its quality depends on the coverage and clustering of the pretraining and adaptation data. Poorly organized or sparse memory entries may lead to suboptimal retrieval, especially for tasks with fine-grained temporal or contact-rich behaviors. Third, cross-domain transfer still requires a small amount of target-domain demonstrations and careful memory-space alignment. The lightweight Bridge-and-Adapt mechanism mitigates representational drift, but it does not guarantee reliable transfer under large embodiment changes, severe visual shifts, or substantially different dynamics. 

\section{Implementation Details}
\label{app:method_details}

This appendix follows the notation and objectives in Secs.~\ref{sec:inline_interface}--\ref{sec:bridge_adapt}. In particular, $A_t$ is the target action chunk, whereas $\mathcal{A}_t$ is a set of active prototype indices. The encoded, retrieved, and adapter-corrected skill representations are denoted by $s_t$, $\hat{s}_t$, and $\tilde{s}_t$, respectively. The implementation details below supplement, rather than redefine, the main-text equations.

\subsection{Notation and Forward Computation}
\label{app:notation_forward}

For each sample, the model receives a non-linguistic observation $O_t=(I_t,x_t)$, comprising visual observations $I_t$ and robot state $x_t$, together with a language instruction $L$. The target action chunk is $A_t=a_{t:t+H-1}\in\mathbb{R}^{H\times d_a}$, where $H$ is the action horizon and $d_a$ is the action dimension. We suppress the batch axis throughout the per-sample equations: $H_t^{\mathrm{vl}}\in\mathbb{R}^{P\times d}$, $E^{\mathrm{mem}}\in\mathbb{R}^{m\times d}$, and $M_t\in\mathbb{R}^{m\times d}$. Here, $P$ is the number of original prefix tokens, $m$ is the number of memory tokens, and $d$ is the hidden dimension. In batched execution, a leading batch dimension of size $B$ is added, and the same learnable memory tokens are broadcast across samples.

The backbone computation is given by Eq.~\eqref{eq:overview_backbone}. The two attention poolers and the Query and Skill Encoders are defined in Eqs.~\eqref{eq:dual_pooling} and~\eqref{eq:query_skill_encode}. For branch $b\in\{q,s\}$, the attention pooler is implemented as
\begin{equation}
    \begin{aligned}
    \beta_{t,i}^{b}
    &=
    \frac{\exp\!\left(u_b^{\top}W_bM_{t,i}^{\top}/\sqrt{d}\right)}
    {\sum_{h=1}^{m}\exp\!\left(u_b^{\top}W_bM_{t,h}^{\top}/\sqrt{d}\right)},
    \qquad i=1,\ldots,m,\\
    \operatorname{AttnPool}_b(M_t)
    &=\sum_{i=1}^{m}\beta_{t,i}^{b}M_{t,i}^{\top}\in\mathbb{R}^{d},
    \end{aligned}
    \label{app:eq:attention_pool}
\end{equation}
where $M_{t,i}\in\mathbb{R}^{1\times d}$ is a memory-token row, $u_b\in\mathbb{R}^{d}$ is a learnable pooling query, and $W_b\in\mathbb{R}^{d\times d}$ is a projection matrix. The transpose in the weighted sum expresses the pooled state as a column vector, consistently with the encoders in Eq.~\eqref{eq:query_skill_encode}. The branch index $b$ is distinct from the Skill Adapter rank $r$ in Eq.~\eqref{eq:skill_adapter}.

The Skill Decoder reconstructs $A_t$ from $s_t$ according to Eq.~\eqref{eq:skill_decoder}; it provides the auxiliary skill-reconstruction supervision in Eq.~\eqref{eq:pretrain_loss} and is not used during deployment. The Token Projector maps a skill representation into policy-conditioning tokens:
\begin{equation}
    T_{\psi}:\mathbb{R}^{d_s}\rightarrow\mathbb{R}^{n_s\times d},
    \qquad
    S_t=
    \begin{cases}
        T_{\psi}(s_t), & \text{Stage A},\\
        T_{\psi}(\tilde{s}_t), & \text{Stages B/C and inference}.
    \end{cases}
    \label{app:eq:token_projector}
\end{equation}
Here, $\tilde{s}_t=A_s(\hat{s}_t)$ is the corrected retrieved skill from Eq.~\eqref{eq:skill_adapter}, with $\tilde{s}_t=\hat{s}_t$ when the adapter is inactive. The Flow Policy predicts $\tilde{A}_t=\pi_{\theta}(H_t^{\mathrm{vl}},S_t)$ as in Eq.~\eqref{eq:overview_action}. The prediction $\tilde{A}_t$ is distinct from the training target $A_t$ and the Skill Decoder reconstruction $\hat{A}_t^{\,s}$.

\subsection{Initial Query-Skill Memory Bank Construction}
\label{app:bank_construction}

After Stage A, we freeze the trained interface and extract query-skill pairs from the pretraining data:
\begin{equation}
    \mathcal{D}_{\mathrm{bank}}
    =\{(q_i,s_i,A_i,\rho_i)\}_{i=1}^{N},
    \label{app:eq:bank_dataset}
\end{equation}
where $i$ indexes a sample, $t_i$ is its within-trajectory time index, and $\rho_i=t_i/T_{\operatorname{traj}(i)}$ is its normalized progress. The denominator $T_{\operatorname{traj}(i)}$ is the duration of the trajectory containing sample $i$. We denote the action chunk associated with this sample by $A_i=(a_{i,0},\ldots,a_{i,H-1})$, where the second subscript indexes an action within that chunk.

\paragraph{Motion-aware downsampling.}
To avoid over-representing static or redundant segments, we compute the action-variation score
\begin{equation}
    \nu_i
    =\frac{1}{H-1}\sum_{h=1}^{H-1}
    \left\|a_{i,h}-a_{i,h-1}\right\|_2,
    \qquad H>1.
    \label{app:eq:motion_score}
\end{equation}
Samples with low $\nu_i$ are downsampled, whereas samples with higher motion variation are retained with higher probability. This indexing keeps action differences within the trajectory and action chunk associated with sample $i$.

\paragraph{Progress-stratified sampling.}
We divide normalized progress into fixed bins and sample from each bin. This prevents the bank from being dominated by a small portion of a trajectory and improves coverage of different execution phases.

\paragraph{Skill clustering.}
We first cluster the retained samples in query space. For a query cluster $\mathcal{C}$, its skill variance is
\begin{equation}
    \sigma_s^2(\mathcal{C})
    =\frac{1}{|\mathcal{C}|}\sum_{i\in\mathcal{C}}
      \left\|s_i-\mu_s(\mathcal{C})\right\|_2^2,
    \qquad
    \mu_s(\mathcal{C})
    =\frac{1}{|\mathcal{C}|}\sum_{i\in\mathcal{C}}s_i.
    \label{app:eq:skill_variance}
\end{equation}
Clusters whose skill variance exceeds the splitting threshold are further partitioned in skill space. Each final cluster $\mathcal{C}_j$ defines an entry of the bank in Eq.~\eqref{eq:memory_bank}:
\begin{equation}
    p_j^q
    =\operatorname{norm}\!\left(
       \frac{1}{|\mathcal{C}_j|}\sum_{i\in\mathcal{C}_j}q_i\right),
    \qquad
    p_j^s=\frac{1}{|\mathcal{C}_j|}\sum_{i\in\mathcal{C}_j}s_i,
    \qquad n_j=|\mathcal{C}_j|.
    \label{app:eq:prototype_init}
\end{equation}
Here, $\operatorname{norm}(v)=v/\|v\|_2$ for a nonzero vector $v$. We retain up to three latent query--skill pairs from each cluster as a compact exemplar set $\mathcal{E}_j$ for replay. The initial bank is denoted by $\mathcal{B}^{(0)}$ and contains $K_0$ entries; the current bank $\mathcal{B}$ contains $K$ entries, which may increase during Stage C.

\subsection{Retrieval and Residual Prototype Adaptation}
\label{app:retrieval_update}

The query alignment $q'_t=A_{\mathrm{mem}}q_t$ is defined in Eq.~\eqref{eq:query_alignment}. For use in both retrieval and the update rules below, let
\begin{equation}
    r_{tj}
    =\operatorname{sim}\!\left(q'_t,p_j^q+\Delta p_j^q\right),
    \qquad j=1,\ldots,K,
    \label{app:eq:retrieval_score}
\end{equation}
where $\operatorname{sim}$ is the same similarity function used in Eq.~\eqref{eq:retrieval_weight}. The set $\mathcal{N}_t$ contains the indices of the top-$K_r$ prototypes under this score. Retrieval weights $\alpha_{tj}$ are computed with temperature $\tau_{\mathrm{mem}}$ by Eq.~\eqref{eq:retrieval_weight}, and the retrieved skill $\hat{s}_t$ is computed by Eq.~\eqref{eq:retrieved_skill}.

For a single sample, $\mathcal{A}_t=\mathcal{N}_t$ is the active index set in Eq.~\eqref{eq:life_loss}. Across a minibatch, the residuals eligible for updates are those retrieved by at least one sample; the per-sample loss retains the active set specified in the main text. Existing base prototypes $(p_j^q,p_j^s)$ remain fixed, while active residuals $(\Delta p_j^q,\Delta p_j^s)$ provide local corrections. Updating support counts and exemplar sets, or appending a new entry in Stage C, is distinct from overwriting an existing base prototype.

\subsection{Bank Update Rule}
\label{app:bank_update_rule}

During Stage C, the reuse decision uses the same residual-adjusted query prototypes as retrieval. For sample $t$, define
\begin{equation}
    r_{t,\max}=\max_{1\leq j\leq K}r_{tj},
    \qquad j_t^{\star}\in\operatorname*{arg\,max}_{1\leq j\leq K}r_{tj}.
    \label{app:eq:max_reuse_similarity}
\end{equation}
If $r_{t,\max}>\tau_{\mathrm{reuse}}$, the sample is assigned to prototype $j_t^{\star}$. It contributes to active-residual learning and updates the corresponding support statistics; its latent query--skill pair is stored as an exemplar, with first-in, first-out replacement once the exemplar set is full. If $r_{t,\max}\leq\tau_{\mathrm{reuse}}$, the sample enters a candidate buffer.

The candidate buffer is periodically clustered. Candidate compactness is evaluated in the same coordinates used to initialize an appended prototype: aligned queries $q'_i$ and encoded skills $s_i$. For a candidate cluster $\mathcal{C}$, define
\begin{equation}
    \begin{aligned}
    \mu_{q'}(\mathcal{C})
      &=\frac{1}{|\mathcal{C}|}\sum_{i\in\mathcal{C}}q'_i,
    &\operatorname{Var}_q(\mathcal{C})
      &=\frac{1}{|\mathcal{C}|}\sum_{i\in\mathcal{C}}
        \|q'_i-\mu_{q'}(\mathcal{C})\|_2^2,\\
    \mu_s(\mathcal{C})
      &=\frac{1}{|\mathcal{C}|}\sum_{i\in\mathcal{C}}s_i,
    &\operatorname{Var}_s(\mathcal{C})
      &=\frac{1}{|\mathcal{C}|}\sum_{i\in\mathcal{C}}
        \|s_i-\mu_s(\mathcal{C})\|_2^2.
    \end{aligned}
    \label{app:eq:candidate_variances}
\end{equation}
A new prototype is appended only when a candidate cluster $\mathcal{C}_{\mathrm{new}}$ satisfies
\begin{equation}
    |\mathcal{C}_{\mathrm{new}}|\geq N_{\min},
    \qquad\operatorname{Var}_q(\mathcal{C}_{\mathrm{new}})\leq\epsilon_q,
    \qquad\operatorname{Var}_s(\mathcal{C}_{\mathrm{new}})\leq\epsilon_s.
    \label{app:eq:new_cluster_condition}
\end{equation}
The appended entry is initialized by
\begin{equation}
    \begin{aligned}
    p_{\mathrm{new}}^q
      &=\operatorname{norm}\!\left(\mu_{q'}(\mathcal{C}_{\mathrm{new}})\right),
    &p_{\mathrm{new}}^s&=\mu_s(\mathcal{C}_{\mathrm{new}}),\\
    n_{\mathrm{new}}&=|\mathcal{C}_{\mathrm{new}}|,
    &\Delta p_{\mathrm{new}}^q&=0,\qquad\Delta p_{\mathrm{new}}^s=0.
    \end{aligned}
    \label{app:eq:new_prototype_init}
\end{equation}
A compact exemplar set $\mathcal{E}_{\mathrm{new}}$ is also stored. Candidate entries use the encoded skills $s_i$; the Skill Adapter operates on the aggregated retrieval $\hat{s}_t$ during policy conditioning and alignment. The adapter, if retained after Stage B, is held fixed during Stage C.

\subsection{Coordinate Alignment under Domain Shift}
\label{app:coordinate_alignment}

Stage B adapts retrieval and skill conditioning under domain shift. The query-side transformation is the linear adapter $A_{\mathrm{mem}}$ in Eq.~\eqref{eq:query_alignment}, initialized close to the identity. For a larger skill-space mismatch, the optional Skill Adapter maps the retrieved skill to $\tilde{s}_t=A_s(\hat{s}_t)$ as defined in Eq.~\eqref{eq:skill_adapter}. Its zero-initialized residual map satisfies $U_sV_s^{\top}=0$, so $\tilde{s}_t=\hat{s}_t$ initially; this condition does not require both factors to be zero. The corrected skill conditions the policy and is aligned to the detached encoded target $s_t$, retaining gradient paths to the adapter through both losses.

Before enabling the Skill Adapter in Stage B, we monitor the novelty rate and retrieval-skill alignment error:
\begin{equation}
    \eta_{\mathrm{batch}}
    =\frac{1}{B}\sum_{i=1}^{B}
      \mathbf{1}\!\left[r_{i,\max}\leq\tau_{\mathrm{reuse}}\right],
    \label{app:eq:novelty_rate}
\end{equation}
\begin{equation}
    e_{\mathrm{align}}
    =\frac{1}{B}\sum_{i=1}^{B}
      \left\|\hat{s}_i-\operatorname{sg}(s_i)\right\|_2^2.
    \label{app:eq:alignment_error}
\end{equation}
Before activation, $\tilde{s}_i=\hat{s}_i$, so this diagnostic measures the unadapted retrieval error. When either signal remains above its activation threshold for several consecutive batches, the adapter is enabled. These diagnostics indicate persistent mismatch but do not, by themselves, distinguish domain shift from genuinely novel skills. During optimization, the encoded target $s_i$ is detached, while the corrected retrieval $\tilde{s}_i$ retains its gradient path.

\subsection{Stage-wise Optimization}
\label{app:stage_optimization}

The three stages use the objectives defined in Sec.~\ref{sec:bridge_adapt}: Stage A minimizes $\mathcal{L}_{\mathrm{pre}}$ in Eq.~\eqref{eq:pretrain_loss}, whereas Stages B and C use the same $\mathcal{L}_{\mathrm{life}}$ in Eq.~\eqref{eq:life_loss}. The stages differ in which parameter groups are updated, not in the definition of the adaptation loss. Table~\ref{tab:trainable_params} summarizes these distinctions. Section~\ref{app:training_objectives} supplies the query-contrastive details and optimization settings without introducing additional stage-specific objective equations.

\section{Training}
\label{app:training_objectives}

\subsection{Training Objectives}

\paragraph{Stage A: interface pretraining.}
Stage A trains the memory seed tokens, both attention poolers, the Query Encoder, the Skill Encoder, the Skill Decoder, the Token Projector, and the last layers of the VLA backbone. No memory bank is used in its forward computation. The policy is conditioned on $S_t=T_{\psi}(s_t)$, and optimization uses $\mathcal{L}_{\mathrm{pre}}$ in Eq.~\eqref{eq:pretrain_loss}, including the original VLA flow-matching loss $\mathcal{L}_{\mathrm{fm}}$ and the Huber reconstruction term $\mathcal{L}_{\mathrm{skill}}$ defined there.

For completeness, we specify the query contrastive term $\mathcal{L}_{q}^{\mathrm{nce}}$ from Eq.~\eqref{eq:pretrain_loss}. Let $\mathcal{J}$ be the set of sample indices in the current minibatch. For sample $i\in\mathcal{J}$, let $y_i$ be its task label, $\operatorname{traj}(i)$ its trajectory identifier, $t_i$ its within-trajectory time index, and $\rho_i$ its normalized progress. The positive set is
\begin{equation}
    \begin{aligned}
    \mathcal{P}(i)
    ={}&\{j\in\mathcal{J}\setminus\{i\}:y_j=y_i,\ |\rho_j-\rho_i|\leq\delta\}\\
    &\cup\{j\in\mathcal{J}\setminus\{i\}:\operatorname{traj}(j)=\operatorname{traj}(i),\ |t_j-t_i|\leq w\}.
    \end{aligned}
    \label{app:eq:query_positives}
\end{equation}
Thus, positives comprise samples from the same task with nearby normalized progress and local neighboring states from the same trajectory, as described in Sec.~\ref{sec:bridge_adapt}. Let $\mathcal{I}=\{i\in\mathcal{J}:|\mathcal{P}(i)|>0\}$ denote the valid anchors. For $\mathcal{I}\neq\varnothing$,
\begin{equation}
    \mathcal{L}_{q}^{\mathrm{nce}}
    =-\frac{1}{|\mathcal{I}|}\sum_{i\in\mathcal{I}}
       \frac{1}{|\mathcal{P}(i)|}\sum_{j\in\mathcal{P}(i)}
       \log\frac{\exp(\operatorname{sim}(q_i,q_j)/\tau_q)}
       {\sum_{\ell\in\mathcal{J}\setminus\{i\}}
        \exp(\operatorname{sim}(q_i,q_{\ell})/\tau_q)}.
    \label{app:eq:query_contrastive_loss}
\end{equation}
When no anchor has a positive, the contrastive contribution is defined to be zero. The query-contrastive temperature $\tau_q$ is distinct from the memory-retrieval temperature $\tau_{\mathrm{mem}}$. After Stage A, the learned interface, Token Projector, Skill Decoder, and backbone are frozen, and the initial memory bank is constructed as described in Sec.~\ref{app:bank_construction}.

\paragraph{Stage B: bridge adaptation.}
Stage B uses $\mathcal{L}_{\mathrm{life}}$ in Eq.~\eqref{eq:life_loss}. It updates the backbone, $A_{\mathrm{mem}}$, $\tau_{\mathrm{mem}}$, and active prototype residuals. The learned interface modules, Token Projector, Skill Decoder, and existing base prototypes remain fixed. Thus, the backbone is made trainable again for this bridge stage, in contrast to its frozen status immediately after Stage A and during Stage C. The optional Skill Adapter follows Eq.~\eqref{eq:skill_adapter} and the activation criterion in Sec.~\ref{app:coordinate_alignment}.

\paragraph{Stage C: memory adaptation.}
Stage C also uses Eq.~\eqref{eq:life_loss}, with the backbone, Inline Memory Interface, Token Projector, and Skill Decoder frozen. It updates only $A_{\mathrm{mem}}$, $\tau_{\mathrm{mem}}$, active prototype residuals, and the bank entries and statistics described in Sec.~\ref{app:bank_update_rule}. The Skill Adapter is not optimized in this stage. For a local batch of $B$ current-task samples, we sample $\lfloor B/3\rfloor$ stored latent query--skill pairs, when available, and combine them with the current samples in the alignment term of Eq.~\eqref{eq:life_loss}, giving an approximate new-to-replay ratio of $3{:}1$. The flow-matching loss is computed only on current-task demonstrations. The replay buffer stores no raw images, actions, or trajectories, and no separate replay loss is introduced.

\begin{table}[!t]
\centering
\caption{Stage-wise optimization and memory-bank operations. ``Last layers'' denotes the partial backbone training in Stage A. The optional Skill Adapter is a Stage-B component; it is fixed or inactive in Stage C.}
\label{tab:trainable_params}
\small
\setlength{\tabcolsep}{5pt}
\begin{tabular}{lccc}
\toprule
\textbf{Parameter group / operation} & \textbf{Stage A} & \textbf{Stage B} & \textbf{Stage C} \\
\midrule
Memory tokens $E^{\mathrm{mem}}$ & train & freeze & freeze \\
Poolers $\operatorname{AttnPool}_q,\operatorname{AttnPool}_s$ & train & freeze & freeze \\
Encoders $f_q,f_s$ & train & freeze & freeze \\
Skill Decoder $D_s$ & train & freeze & freeze \\
Token Projector $T_{\psi}$ & train & freeze & freeze \\
Backbone $F_{\theta}$ & last layers & train & freeze \\
Query alignment $A_{\mathrm{mem}}$ & -- & train & train \\
Retrieval temperature $\tau_{\mathrm{mem}}$ & -- & train & train \\
Active residuals $\Delta p_j^q,\Delta p_j^s$ & -- & train & train \\
Skill Adapter $A_s$ & -- & train & freeze \\
Bank structure $\mathcal{B}$ & -- & -- & update \\
\bottomrule
\end{tabular}
\end{table}

\subsection{Hyperparameter Settings}
\label{app:hyperparameters}

Table~\ref{tab:default_hyperparameters} summarizes the hyperparameters reported for Optimus-R. Unless otherwise specified, the same configuration is used across experiments. The symbols match the main text: $K_r$ is the number of retrieved prototypes, $K_0$ is the initial bank size, and $r$ is the rank of the optional Skill Adapter.

\begin{table}[!t]
\centering
\caption{Reported default hyperparameters for Optimus-R.}
\label{tab:default_hyperparameters}
\small
\setlength{\tabcolsep}{5pt}
\begin{tabular}{llc}
\toprule
\textbf{Category} & \textbf{Hyperparameter} & \textbf{Value} \\
\midrule
\multirow{4}{*}{Architecture}
& Memory tokens $m$ & $4$ \\
& Query dimension $d_q$ & $256$ \\
& Skill dimension $d_s$ & $256$ \\
& Skill tokens $n_s$ & $4$ \\
\midrule
\multirow{4}{*}{Memory bank / retrieval}
& Initial bank size $K_0$ & $128$ \\
& Retrieval count $K_r$ & $2$ \\
& Reuse threshold $\tau_{\mathrm{reuse}}$ & $0.75$ \\
& Minimum cluster size $N_{\min}$ & $50$ \\
\midrule
\multirow{8}{*}{Stage A pretraining}
& Query contrastive weight $\lambda_q$ & $0.1$ \\
& Skill reconstruction weight $\lambda_s$ & $1.0$ \\
& Query contrastive temperature $\tau_q$ & $0.07$ \\
& Progress threshold $\delta$ & $0.1$ \\
& Trajectory window $w$ & $6$ \\
& Learning rate for new modules & $2\times10^{-4}$ \\
& Learning rate for unfrozen backbone layers & $2\times10^{-5}$ \\
& Weight decay for new modules & $10^{-4}$ \\
\midrule
\multirow{5}{*}{Stage B/C adaptation}
& Alignment weight $\lambda_{\mathrm{align}}$ & $0.1$ \\
& Residual regularization weight $\lambda_{\mathrm{reg}}$ & $10^{-4}$ \\
& Learning rate for $A_{\mathrm{mem}}$ & $10^{-4}$ \\
& Learning rate for $\tau_{\mathrm{mem}}$ & $5\times10^{-4}$ \\
& Learning rate for active prototype residuals & $10^{-4}$ \\
\midrule
\multirow{2}{*}{Optional Skill Adapter}
& Adapter rank $r$ & $8$ \\
& Reported learning rate for $A_s$ & $5\times10^{-5}$ \\
\bottomrule
\end{tabular}
\end{table}

For Stage A, the newly introduced modules use a learning rate of $2\times10^{-4}$ and weight decay $10^{-4}$, while the unfrozen backbone layers use a learning rate of $2\times10^{-5}$. For both adaptation stages, $\lambda_{\mathrm{align}}=0.1$ and $\lambda_{\mathrm{reg}}=10^{-4}$. The backbone is updated in Stage B and frozen in Stage C; the learned interface modules and Token Projector remain fixed in both. The optional residual Skill Adapter has rank $r=8$, and the reported adapter learning rate is $5\times10^{-5}$; this setting does not imply that the adapter is optimized in Stage C.

\subsection{Inference}
\label{app:inference}

At inference time, neither the Skill Encoder nor the Skill Decoder is needed. Given $(O_t,L)$, the model computes the memory-token states by Eq.~\eqref{eq:overview_backbone} and forms the retrieval query with the query branch of Eqs.~\eqref{eq:dual_pooling} and~\eqref{eq:query_skill_encode}. It then aligns $q_t$ to obtain $q'_t=A_{\mathrm{mem}}q_t$, selects $\mathcal{N}_t$, and computes $\alpha_{tj}$ and $\hat{s}_t$ according to Eqs.~\eqref{eq:query_alignment}--\eqref{eq:retrieved_skill}. This top-$K_r$ retrieval determines what to retrieve; no binary retrieval gate is used. If enabled, the Skill Adapter produces $\tilde{s}_t=A_s(\hat{s}_t)$; otherwise, $\tilde{s}_t=\hat{s}_t$. Finally, $S_t=T_{\psi}(\tilde{s}_t)$ conditions the Flow Policy, which predicts $\tilde{A}_t=\pi_{\theta}(H_t^{\mathrm{vl}},S_t)$ as in Eq.~\eqref{eq:overview_action}.

In a cache-based implementation, the original multimodal and memory-token prefix is cached and then extended with $S_t$ before action generation. This is an implementation of the same policy-conditioning path, not a separate backbone or action generator. The skill-side attention pooler, Skill Encoder, and Skill Decoder are unnecessary for this retrieval-only inference path; the learned query alignment, retrieval temperature, prototype residuals, and any enabled Skill Adapter are retained.

\section{Evaluation on RoboTwin 2.0}
RoboTwin~2.0 provides a standardized benchmark for bimanual manipulation built upon the RoboTwin-OD object library, which contains $731$ annotated objects spanning $147$ categories, together with a large-scale corpus of more than $100\text{k}$ expert dual-arm trajectories. The benchmark includes $50$ collaborative bimanual tasks instantiated across five robot embodiments. Under the standard simulation protocol, each task is trained independently on the Aloha–AgileX dual-arm platform using $50$ clean expert demonstrations, and evaluated with $100$ rollouts under two difficulty settings: an \emph{Easy} setting with uncluttered scenes, and a \emph{Hard} setting with substantial domain randomization, including clutter, background textures, lighting variations, and tabletop-height perturbations.

\begin{table*}[!t]
\centering
\renewcommand{\arraystretch}{1.1}
\caption{Performance comparison on RoboTwin 2.0. We report per-task success rates (SR) over 100 rollouts under \textit{Hard} setting. $^{\dag}$ represents the result we reproduced.}
\label{tab:exp_robotwin_all}
\small
\setlength{\tabcolsep}{4pt}

\resizebox{0.65\linewidth}{!}{
\begin{tabular}{l|ccccccc}
\toprule[1.2pt]
\textbf{Task} 
& \textbf{RDT}
& \textbf{ACT}
& \textbf{DP}
& \textbf{DP3}
& $\mathbf{\pi_{0}}$
& $\mathbf{\pi_{0.5}}^{\dag}$
& \textbf{Optimus-R} \\
\midrule
Adjust Bottle
& 75\% & 23\% & 0\% & 3\% & 56\% & 75\% & \textbf{89\%} \\
Click Alarmclock
& 12\% & 4\% & 5\% & 14\% & 11\% & \textbf{44\%} & 41\% \\
Click Bell
& 9\% & 3\% & 0\% & 0\% & 3\% & \textbf{64\%} & 53\% \\
Grab Roller
& 43\% & 25\% & 0\% & 2\% & 80\% & 82\% & \textbf{94\%} \\
Move Playingcard Away
& 11\% & 0\% & 0\% & 3\% & 22\% & 32\% & \textbf{38\%} \\
Pick Diverse Bottles
& 0\% & 0\% & 0\% & 1\% & 6\% & 29\% & \textbf{36\%} \\
Place a2b Left
& 1\% & 0\% & 0\% & 2\% & 1\% & 20\% & \textbf{29\%} \\
Place a2b Right
& 1\% & 0\% & 0\% & 0\% & 6\% & 19\% & \textbf{21\%} \\
Place Bread Basket
& 2\% & 0\% & 0\% & 1\% & 4\% & 28\% & \textbf{39\%} \\
Place Burger Fries
& 27\% & 0\% & 0\% & 18\% & 4\% & 46\% & \textbf{59\%} \\
Place Container Plate
& 17\% & 1\% & 0\% & 1\% & 45\% & 55\% & \textbf{74\%} \\
Place Empty Cup
& 7\% & 0\% & 0\% & 1\% & 11\% & 59\% & \textbf{72\%} \\
Place Object Stand
& 5\% & 0\% & 0\% & 0\% & 11\% & 46\% & \textbf{47\%} \\
Place Shoe
& 7\% & 0\% & 0\% & 2\% & 6\% & 20\% & \textbf{40\%} \\
Rotate QRCode
& 5\% & 0\% & 0\% & 1\% & 15\% & \textbf{20\%} & \textbf{20\%} \\
Shake Bottle Horizon
& 51\% & 4\% & 18\% & 25\% & 51\% & 85\% & \textbf{94\%} \\
Shake Bottle
& 45\% & 10\% & 8\% & 19\% & 60\% & 82\% & \textbf{96\%} \\
Stack Blocks Two
& 2\% & 0\% & 0\% & 0\% & 1\% & 21\% & \textbf{35\%} \\
Stack Bowls Two
& 30\% & 0\% & 0\% & 6\% & 41\% & 62\% & \textbf{71\%} \\
Stack Bowls Three
& 17\% & 0\% & 0\% & 5\% & 24\% & 42\% & \textbf{51\%} \\
\bottomrule[1.2pt]
\end{tabular}
}
\end{table*}

In our experiments, we evaluate on a randomly selected subset of $20$ tasks. For each task, we train our models using all available clean demonstrations and report performance over $100$ rollouts in the \emph{Hard} setting, which provides a stringent test of robustness to visual and physical domain shifts. As shown in Table~\ref{tab:exp_robotwin_all}, Optimus-R obtains an average success rate of $55\%$ across the selected tasks, outperforming all compared baselines, including $\pi_{0.5}$.

\section{Real-World Setting}
\paragraph{Sim-to-real adaptation.}
\label{app:sim2real_setting}
In the transfer setting of Sec.~\ref{subsec:sim2real}, the RoboTwin and real-world data use a common observation format and action-space representation. The target domain differs in backgrounds, lighting, object appearance, camera viewpoints, and spatial layouts. For example, the real-world \textit{Block Stacking} task shares manipulation patterns with RoboTwin's \textit{Stack Blocks Two}, despite different visual conditions. During Stage B, $A_{\mathrm{mem}}$ maps target-domain queries into the source memory coordinates, as defined in Eq.~\eqref{eq:query_alignment}, so that relevant source skills can be retrieved and refined through prototype residual updates.

To rigorously evaluate the lifelong learning capability and the anti-forgetting mechanisms of Optimus-R, we designed a comprehensive suite of real-world robotic manipulation tasks. The evaluation protocol is specifically structured to simulate a challenging continuous learning scenario, encompassing a transition from basic single-arm manipulations to complex, long-horizon bimanual coordination. 

\begin{figure*}[!t]
    \centering
    \includegraphics[width=1\textwidth]{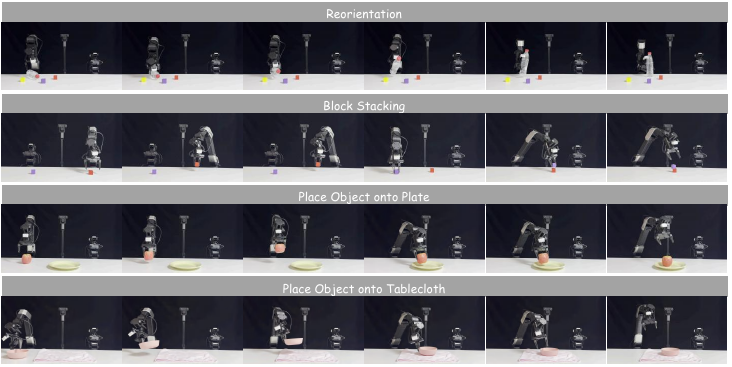}
    \caption{
    Qualitative rollouts on real-world Category-A tasks. 
    We visualize representative trajectories for four previously learned manipulation tasks: 
    \emph{Reorientation}, \emph{Block Stacking}, \emph{Place Object onto Plate}, and 
    \emph{Place Object onto Tablecloth}. 
    Across these tasks, Optimus-R consistently executes precise object-centric manipulation behaviors 
    under diverse object and target configurations. 
    These results illustrate that the learned query-skill memory can preserve reusable visuomotor 
    primitives and support stable execution of previously acquired skills during subsequent adaptation.
    }
    \label{fig:appendix-case1}
\end{figure*}

\begin{figure*}[!t]
    \centering
    \includegraphics[width=1\textwidth]{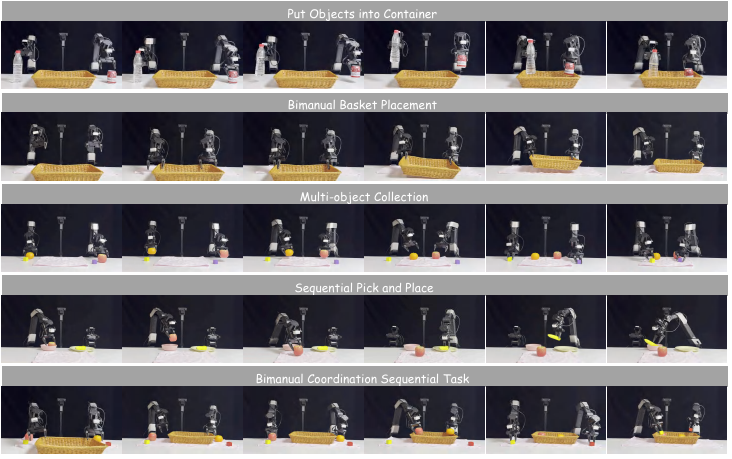}
    \caption{
    Qualitative rollouts on real-world Category-B tasks under the lifelong adaptation setting. 
    We show representative trajectories for five newly introduced tasks: 
    \emph{Put Objects into Container}, \emph{Bimanual Basket Placement}, 
    \emph{Multi-object Collection}, \emph{Sequential Pick and Place}, and 
    \emph{Bimanual Coordination Sequential Task}. 
    These tasks require multi-object reasoning, long-horizon sequencing, and coordinated dual-arm control, 
    posing a substantial distribution shift from the Category-A task set. 
    Optimus-R successfully adapts to these novel behaviors by retrieving, updating, and expanding 
    explicit query-skill memory entries, enabling new skill acquisition while mitigating interference 
    with previously learned manipulation capabilities.
    }
    \label{fig:appendix-case2}
\end{figure*}

As summarized in Table~\ref{tab:real_world_tasks}, the tasks are strategically divided into two distinct categories:
\begin{itemize}[leftmargin=*]
    \item \textbf{Category A (Base Tasks):} This set comprises 4 fundamental single-arm manipulation tasks(Figs. \ref{fig:appendix-case1}), including spatial reorientation and precision placement (e.g., block stacking, placing objects onto specific receptacles). The model is initially trained on this task suite, which provides a total of 570 demonstrations. These tasks serve as the foundation for evaluating the model's stability (i.e., retention of historical skills) during subsequent lifelong adaptation.
    \item \textbf{Category B (Novel Tasks):} After acquiring the base skills, the model is incrementally adapted to a stream of 5 novel tasks(Figs.\ref{fig:appendix-case2}). Crucially, these tasks introduce a significant shift in the control distribution by requiring \emph{bimanual coordination} and \emph{long-horizon sequential logic} (e.g., sequential pick-and-place, multi-object collection). This category contains 420 demonstrations in total.
\end{itemize}

This Category A $\rightarrow$ B transition is intentionally demanding. The shift from single-arm to bimanual control, coupled with the introduction of multi-stage semantic reasoning, typically exacerbates catastrophic forgetting in parameter-centric fine-tuning methods. Evaluating Optimus-R under this protocol effectively demonstrates the robustness of our discrete Query-Skill Memory Bank in preserving old capabilities while assimilating highly distinct new skills.

\begin{table}[!t]
\centering
\caption{Detailed configuration of the real-world tasks for lifelong learning evaluation. The tasks are divided into Base Tasks (Category A) and Novel Tasks (Category B), featuring a transition from single-arm to bimanual manipulation.}
\label{tab:real_world_tasks}
\resizebox{\textwidth}{!}{
\begin{tabular}{@{}c p{6cm} c p{3.5cm} p{3.5cm}@{}}
\toprule
\textbf{Cat.} & \textbf{Task Family} & \textbf{Arm Mode} & \textbf{Objects} & \textbf{Targets} \\ 
\midrule
\multirow{4}{*}{A} 
& Reorientation & Single-arm & bottle, mug & - \\ \cmidrule(l){2-5} 
& Block Stacking & Single-arm & block & - \\ \cmidrule(l){2-5} 
& Place Object onto Plate & Single-arm & can, bowl, fruit & plate \\ \cmidrule(l){2-5} 
& Place Object onto Tablecloth & Single-arm & bowl, plate, cup & tablecloth \\ 
\midrule
\multirow{5}{*}{B} 
& Put Objects into Container & Bimanual & bottle, can & basket \\ \cmidrule(l){2-5} 
& Bimanual Basket Placement & Bimanual & basket & tablecloth \\ \cmidrule(l){2-5} 
& Multi-object Collection & Bimanual & fruit, block & plate, tablecloth \\ \cmidrule(l){2-5} 
& Sequential Pick and Place & Bimanual & fruit, cup & bowl, tablecloth, plate \\ \cmidrule(l){2-5} 
& Bimanual Coordination Sequential Task & Bimanual & basket, fruits & tablecloth, basket \\ 
\bottomrule
\end{tabular}
}
\end{table}


\clearpage
\section{Additional Experiments}
\label{app:additional_experiments}

\subsection{Memory-Bank Component Ablations}
\label{app:bank_component_ablation}

We evaluate memory-bank design choices on LIBERO using 100\% of the training demonstrations and report the average success rate across the four suites. Table~\ref{tab:app_bank_ablation} compares the full model with variants that omit skill-aware clustering, omit exemplar storage for replay, or replace motion-aware downsampling with uniform downsampling.

\begin{table}[!t]
\centering
\small
\caption{Memory-bank component ablations on LIBERO with 100\% training data.}
\label{tab:app_bank_ablation}
\setlength{\tabcolsep}{8pt}
\begin{tabular}{l|c}
\toprule[1.2pt]
\textbf{Variant} & \textbf{Average SR (\%)} \\
\midrule
\rowcolor[HTML]{E7EEFE}
Optimus-R (Full) & \textbf{97.8} \\
w/o skill-aware clustering & 95.8 \\
w/o exemplar storage & 96.3 \\
Uniform downsampling & 97.2 \\
\bottomrule[1.2pt]
\end{tabular}
\end{table}

Removing clustering or exemplar storage reduces success by 2.0 and 1.5 percentage points, respectively, supporting their contribution to memory-based adaptation. Uniform downsampling produces a smaller decrease of 0.6 percentage points, suggesting that motion-aware sampling is beneficial but less influential among the tested variants.

\subsection{Memory-Bank Configuration Sensitivity}
\label{app:bank_sensitivity}

We examine the initial bank size $K_0$, retrieval count $K_r$, and number of memory tokens $m$ on LIBERO at 100\% training data. Table~\ref{tab:app_bank_sensitivity} distinguishes changes to a single parameter from configurations that jointly change $K_r$ and $m$ relative to the default $(128,2,4)$.

\begin{table}[!t]
\centering
\small
\caption{LIBERO configuration sensitivity. Bold entries identify the default configuration. The last two rows jointly vary retrieval count and memory-token count.}
\label{tab:app_bank_sensitivity}
\setlength{\tabcolsep}{10pt}
\begin{tabular}{ccc|c}
\toprule[1.2pt]
$\boldsymbol{K_0}$ & $\boldsymbol{K_r}$ & $\boldsymbol{m}$ & \textbf{Average SR (\%)} \\
\midrule
64 & 2 & 4 & 97.2 \\
\rowcolor[HTML]{E7EEFE}
\textbf{128} & \textbf{2} & \textbf{4} & \textbf{97.8} \\
256 & 2 & 4 & 97.8 \\
128 & 1 & 4 & 97.0 \\
\midrule
128 & 4 & 8 & 97.4 \\
128 & 4 & 1 & 96.8 \\
\bottomrule[1.2pt]
\end{tabular}
\end{table}

Success rates range from 96.8\% to 97.8\% across the tested configurations. Increasing $K_0$ from 128 to 256 yields no further gain, while reducing $K_r$ from two to one lowers success to 97.0\%. The default therefore attains the highest observed success with fewer prototypes than the 256-entry bank. The joint configurations do not isolate the individual effects of $K_r$ and $m$ relative to the default.

\subsection{Memory Growth across Task Suites}
\label{app:memory_growth}

To examine memory growth over a longer adaptation sequence, we initialize Optimus-R on LIBERO-90 and then adapt sequentially to Spatial, Object, Goal, and Long. Table~\ref{tab:app_memory_growth} reports the occupied prototype count $K$, the increment at each stage, retrieval latency, and Recall@5. This sequence starts with 81 occupied prototypes; the counts describe the evolving bank in this experiment rather than the default $K_0=128$ configuration above.

\begin{table}[!t]
\centering
\small
\caption{Memory growth during sequential suite adaptation. Prototype increments are relative to the preceding stage; latency measures retrieval rather than full policy inference.}
\label{tab:app_memory_growth}
\setlength{\tabcolsep}{6pt}
\begin{tabular}{l|cccc}
\toprule[1.2pt]
\textbf{Adaptation stage} & $\boldsymbol{K}$ & \textbf{New} & \textbf{Latency (ms)} & \textbf{Recall@5 (\%)} \\
\midrule
After LIBERO-90 & 81 & -- & 1.80 & 95.6 \\
$+$ Spatial & 84 & 3 & 1.80 & 95.4 \\
$+$ Object & 90 & 6 & 1.83 & 96.1 \\
$+$ Goal & 99 & 9 & 1.86 & 95.2 \\
\rowcolor[HTML]{E7EEFE}
$+$ Long & 109 & 10 & 1.90 & 94.7 \\
\bottomrule[1.2pt]
\end{tabular}
\end{table}

The bank adds 28 prototypes across four suites, while retrieval latency increases by 0.10\,ms and Recall@5 changes from 95.6\% to 94.7\%. These measurements indicate a modest retrieval overhead over the evaluated sequence. They characterize expansion at this scale, without establishing an upper bound on memory growth or prototype interference.

\subsection{Cross-Suite Transfer from Spatial to Goal}
\label{app:spatial_goal_transfer}

Starting from the $\pi_{0.5}$-base checkpoint, we train Optimus-R on the ten LIBERO-Spatial tasks with $K_0=16$, evaluate it zero-shot on LIBERO-Goal, and then adapt it to Goal. The suites differ in scene layouts and initial-state distributions while sharing manipulation patterns, such as placing a bowl on a plate. Table~\ref{tab:app_spatial_goal} reports overall Goal performance, performance on this shared manipulation pattern, and bank size.

\begin{table}[!t]
\centering
\small
\caption{Spatial-to-Goal transfer. Bowl-on-plate denotes the Goal task \textit{put the bowl on the plate}.}
\label{tab:app_spatial_goal}
\setlength{\tabcolsep}{7pt}
\begin{tabular}{l|ccc}
\toprule[1.2pt]
\textbf{Evaluation stage} & \textbf{Goal SR (\%)} & \textbf{Bowl-on-plate SR (\%)} & $\boldsymbol{K}$ \\
\midrule
Zero-shot & 8 & 68 & 16 \\
\rowcolor[HTML]{E7EEFE}
After Goal adaptation & \textbf{96} & \textbf{98} & 27 \\
\bottomrule[1.2pt]
\end{tabular}
\end{table}

The bowl-on-plate task achieves 68\% success before target-suite adaptation, consistent with transfer of a manipulation pattern represented in Spatial. However, overall zero-shot success is only 8\%, and \textit{turn on the stove} provides a failure case involving an operation absent from the Spatial tasks. After adaptation, overall success reaches 96\% and the bank grows by 11 prototypes. The results support selective transfer of shared behaviors while showing the need for target-task adaptation.

\subsection{Task-Wise Continual Learning on LIBERO-Spatial}
\label{app:spatial_continual}

We further evaluate a ten-task sequence on LIBERO-Spatial using the task order of \citet{liu2026resistant}. Optimus-R starts from $\pi_{0.5}$-base and trains for 10,000 steps per task, carrying the preceding checkpoint forward. After each stage, we evaluate all tasks observed so far. Adaptation uses the latent exemplar replay described in Sec.~\ref{app:training_objectives}. Table~\ref{tab:app_spatial_continual} reports the resulting success-rate matrix.

\begin{table}[!t]
\centering
\small
\caption{Task-wise continual learning on LIBERO-Spatial. Entries are success rates (\%); dashes denote tasks not yet introduced. $T_1$--$T_{10}$ follow the sequence in \citet{liu2026resistant}.}
\label{tab:app_spatial_continual}
\setlength{\tabcolsep}{5pt}
\begin{tabular}{l|cccccccccc}
\toprule[1.2pt]
\textbf{Checkpoint} & $\boldsymbol{T_1}$ & $\boldsymbol{T_2}$ & $\boldsymbol{T_3}$ & $\boldsymbol{T_4}$ & $\boldsymbol{T_5}$ & $\boldsymbol{T_6}$ & $\boldsymbol{T_7}$ & $\boldsymbol{T_8}$ & $\boldsymbol{T_9}$ & $\boldsymbol{T_{10}}$ \\
\midrule
After $T_1$ & 100 & -- & -- & -- & -- & -- & -- & -- & -- & -- \\
After $T_2$ & 98 & 100 & -- & -- & -- & -- & -- & -- & -- & -- \\
After $T_3$ & 98 & 100 & 100 & -- & -- & -- & -- & -- & -- & -- \\
After $T_4$ & 100 & 98 & 100 & 98 & -- & -- & -- & -- & -- & -- \\
After $T_5$ & 100 & 100 & 94 & 100 & 98 & -- & -- & -- & -- & -- \\
After $T_6$ & 96 & 100 & 96 & 98 & 94 & 98 & -- & -- & -- & -- \\
After $T_7$ & 98 & 100 & 96 & 98 & 94 & 96 & 100 & -- & -- & -- \\
After $T_8$ & 98 & 100 & 100 & 100 & 92 & 94 & 98 & 98 & -- & -- \\
After $T_9$ & 100 & 98 & 98 & 100 & 92 & 100 & 100 & 98 & 98 & -- \\
\rowcolor[HTML]{E7EEFE}
After $T_{10}$ & 100 & 98 & 100 & 98 & 90 & 98 & 98 & 98 & 100 & 96 \\
\bottomrule[1.2pt]
\end{tabular}
\end{table}

We also report negative backward transfer (NBT): for each task, we average its success-rate decrease from its post-training checkpoint over subsequent stages, then average over all ten tasks, assigning zero to the final task. Using success rates on a $[0,1]$ scale, the sequence yields NBT $=0.009$ and a final average success rate of 97.6\%. This indicates limited average forgetting over the sequence, although retention varies by task; for example, $T_5$ decreases from 98\% immediately after training to 90\% at the final stage.

\end{document}